\documentclass[]{interact}

\usepackage{algorithm} 
\usepackage{algpseudocode} 
\usepackage{epstopdf}% To incorporate .eps illustrations using PDFLaTeX, etc.
\usepackage{subfigure}% Support for small, `sub' figures and tables
\usepackage{xcolor}
\usepackage{listings}
\usepackage{url}
\usepackage{ulem}
\usepackage{textcomp}
\usepackage{setspace}

\newcommand{\C}{\mathbf{C}}
\newcommand{\LL}{\mathbf{L}}

\newcommand{\X}{\mathbf{X}}

\newcommand{\Xtrans}{\mathbf{Z}}

\newcommand{\xvec}{\mathbf{x}} %\x seems to perhaps by defined by the TnF style file?
\newcommand{\y}{\mathbf{y}}
\newcommand{\K}{\mathbf{K}}

\newcommand*{\Esp}[2]{\mathbb{E}_{#1}\left[#2\right]}

\newcommand*{\Var}[2]{\mathbb{V}\mathrm{ar}_{#1}\left[#2\right]}

\newcommand{\xnew}{\tilde{\xvec}}

\newcommand{\nvar}{p}
\newcommand{\nsamp}{n}
\newcommand{\nred}{r}

\usepackage{natbib}
\theoremstyle{plain}% Theorem-like structures

\theoremstyle{definition}

\theoremstyle{remark}

\def\spacingset#1{\renewcommand{\baselinestretch}%
{#1}\small\normalsize} \spacingset{1}

\begin{document}

\articletype{Preprint}

\title{Sensitivity Prewarping for Local Surrogate Modeling}

\author{
\name{Nathan Wycoff\textsuperscript{1}\thanks{To whom correspondence should be addressed: Nathan Wycoff (nathan.wycoff@georgetown.edu)}, Mickaël Binois\textsuperscript{2} and Robert B. Gramacy\textsuperscript{3}}
\affil{\textsuperscript{1} McCourt School
of Public Policy, Georgetown University; \textsuperscript{2} ACUMES, Inria Sophia Antipolis; \textsuperscript{3} Dept. of Statistics, Virginia Tech}
}
%\author{Anonymous}

\maketitle

\begin{abstract}
%However, vanilla GPs scale poorly with sample size and make potentially unrealistic prior stationarity assumptions on the simulator, problems which are both solved via local surrogate modeling.
In the continual effort to improve product quality and decrease operations costs, computational modeling is increasingly being deployed to determine feasibility of product designs or configurations. Surrogate modeling of these computer experiments via local models, which induce sparsity by only considering short range interactions, can tackle huge analyses of complicated input-output relationships. However, narrowing focus to local scale means that global trends must be re-learned over and over again. 
In this article, we propose a framework for incorporating information from a global sensitivity analysis into the surrogate model as an input rotation and rescaling preprocessing step. We discuss the relationship between several sensitivity analysis methods based on kernel regression before describing how they give rise to a transformation of the input variables. 
%a subbagging estimator we prefer in our large-data surrogate modeling context.
%We show that this preprocessing scheme frees local models to better focus on local information in the sense that the transformed simulator is equally sensitive to all input directions.
Specifically, we perform an input warping such that the ``warped simulator'' is equally sensitive to all input directions, freeing local models to focus on local dynamics.
Numerical experiments on observational data and benchmark test functions, including a high-dimensional computer simulator from the automotive industry, provide empirical validation.
\end{abstract}

\begin{keywords}
computer experiments; emulation; sensitivity analysis; Gaussian process; dimension reduction; active subspace; subbagging
    \end{keywords}
    
\newpage
\spacingset{1.8} % DON'T change the spacing!

\section{Introduction}

As previously unimaginable computing power has become widely available, industrial scientists are increasingly making use of computationally intensive computer programs to simulate complex phenomena that cannot be explained by simple mathematical models and which would be prohibitively expensive to experiment upon physically. 
These computer experiments have varied business applications, for example:
\cite{Zhou2013} describe virtualization of an injection molding process; \cite{Montgomery2001} explored the strength of automobile components; \cite{Crema2015} developed a computer model to help manage an assemble to order system.  Despite the tremendous supply of computational resources provided by increasingly powerful CPUs, the general purpose GPU computing paradigm, and even more specialized hardware such as tensor processing units, the demands of advanced computer models are still sizeable. As such, there is a market for fitting surrogates to computer simulations: flexible statistical models which learn the input-output mapping defined by the simulator of interest, and are ideally suited to serve as a substitute for the same.   For detailed review, see \cite{Gramacy2020surrogates,santner2018design, Forrester2008}.

One popular use of computer experiments is to perform  sensitivity analysis \cite[e.g.,][Ch.~8.2]{oakley2004probabilistic,marrel2009calculations,gramacy2013variable,da2009local,Gramacy2020surrogates}.
This can consist of determining which of the input parameters are most influential, or even whether some latent combination of the inputs is  driving the response.
Sensitivity analysis for computer experiments must take into account unique characteristics not found in observational data.
As in classical design of experiments, the training data inputs can be chosen, which means there is no need to take into account natural correlation between the input variables.  Moreover, the design may be selected to maximize information gain or other criteria \cite[][Ch.~6]{Gramacy2020surrogates}.  
Further, in the case of deterministic experiments, we observe input-output dynamics exactly, and sometimes may even have derivative information (or can approximate it).
Active Subspaces \cite[AS;][]{Constantine2015} exploit the knowledge of these gradients to perform linear sensitivity analysis, that is to say, sensitivity analysis which finds ``directions", or linear combinations of inputs, of greatest influence, rather than evaluating individual input variables.
In this article, we will not assume knowledge of the gradient, but we will leverage that the target simulator is smooth, such that we can estimate its AS nonparametrically \citep{Othmer2016,palar2017,Palar2018,sasl}.
These methods are closely related to existing gradient-based kernel dimension reduction \citep{Fukumizu2014} techniques from the statistics literature, which we discuss in a unified framework in Section \ref{sec:gbsa}. 

Global Sensitivity Analysis (GSA), beyond being of interest in and of itself, can also be used to perform a transformation to the input space before applying standard modeling methods, a process referred to as premodeling in \cite{li2005}. Sometimes, this can take the form of variable selection, as in using lasso to select variables before fitting a standard linear model \citep{Belloni2013}. Otherwise, the dimension of the space is not changed, but simply our orientation within it, for instance by changing basis to that implied by Principal Components Analysis (PCA). This has been recommended as a preprocessor for ``axis-aligned" methods such as generalized additive models \citep{deSouza2017} and tree-based methods \citep{Rodriguez2006}. And, of course, these approaches can be combined to learn both a rotated and truncated space, as in principal components regression \citep{HastieElements}. 

In this article, we argue that this approach also has much promise as a preprocessor for local surrogate modeling of large-scale computer experiments \citep[e.g.,][]{Gramacy2015lagp,Katzfuss2020}. Practically speaking, what we dub ``prewarping'' influences the local model both directly and indirectly. Directly, because it redefines the definition of distances between points upon which many surrogate models (e.g., those based on Gaussian process regression) rely to compute relevant spatial correlations, and indirectly, as the definition of ``local'' changes with the metric, thus influencing neighborhood selection. We build on recently proposed linear GSA techniques and show significant improvement compared to directly applying the local methods to the original input space. Intuitively, GSA based preprocessing handles global trends, and frees the local models to better represent nearby information. 
We formalize this intuition in Section \ref{sec:methtrans} by proposing that the relationship between the warped inputs and the outputs be equally sensitive to every input dimension at the global level. We find that this enhances predictive ability on a battery of test functions and datasets. 

This prewarping idea may be compared to preconditioning in numerical analysis \citep{Wathen2015},
where a central problem is the solution of linear systems $\mathbf{A}\mathbf{x}=\mathbf{b}$.  Modern solution algorithms are typically iterative, meaning that they operate by improving a given approximate solution $\tilde{\mathbf{x}}$ over the course of many iterations until a measure of error like $||\mathbf{A}\tilde{\mathbf{x}}-\mathbf{b}||$ is acceptable. 
%Numerical analysis researchers have found that oftentimes, they can get away with using fewer iterations by first performing a linear transformation to the input space  to improve the conditioning of the linear system; this technique is called preconditioning \cite{Wathen2015}.
Numerical analysts have found that oftentimes, by first performing a linear transformation to the input space, they improve the conditioning of the linear system which results in fewer iterations required for a given level of accuracy.
Similarly, we propose performing a linear transformation of the input space based on a GSA in the hope that this will result in fewer \textit{data} requirements for a given level of accuracy, or greater accuracy given data.
If a surrogate prior to the linear transformation corresponds to fitting $y_i$ versus $\xvec_i$, afterwards the problem becomes $y_i$ versus $\LL\xvec_i$, where $\LL$ is derived from an appropriate GSA. %\textbf{}\change{Ensure consistency of acronyms, abbevations, and capitalizations w/ journal}.

In particular, given a large collection of simulator inputs $\X$ and outputs $\y$, we propose first conducting a GSA using a Gaussian Process (GP) fit to a (global) manageably-sized subset of the data.  We prefer a separable kernel (details in Section \ref{sec:bg}), learning correlation decay separately along each dimension.  The Automatic Relevance Determination  \citep[ARD;][Ch.~5.1]{Neal1996,Rasmussen2006} principle holds that those input dimensions with large kernel length-scales are less important, and can be dropped when conducting variable selection.
Scaling each input dimension by the reciprocal of the associated length-scale, one possible $\LL$, thus imbues the local surrogate with inductive bias reflecting global trends.  

PCA is an option that goes beyond re-scaling to linear projection.
% rotated regression trees do more than (not sure introducing trees is helpful)
% rescale the existing variables, they essentially form new ones via linear combination.
However, PCA's emphasis on dispersion means it's less useful for surrogate modeling, where designs are typically chosen by the practitioner;  i.e., no input dispersion to learn beyond that we have ourselves imposed.
%Though PCA is at teasing out the directions across which the input data are dispersed, 
%In the context of regression on observational data, PCA is useful in that it will emphasize directions which the input data are dispersed across.
%However, the fact that it does not even consider the response $\y$ leaves something to be desired.
%In the context of surrogate modeling, however, PCA is even less informative, since as the design points are chosen by the practitioner, there is no input dispersion to learn beyond that which we have ourselves imposed.
AS, however, allows for non-axis aligned measures of sensitivity, emitting an $\LL$ for the purposes of linear projection, while also accounting for the response.   We provide the details of how such sensitivities may be realized through efficient sampling schemes, and how $\LL$ may be backed out for the purposes of input warping for downstream local analysis, and ultimately accurate prediction.  We privilege AS $\LL$ prewarping as well as two axis-by-axis sensitivity analyses, which we show both improve upon simple global and local schemes, however there are certainly other possibilities.

After reviewing relevant background in Section \ref{sec:bg}, our proposed methodology is detailed in Section \ref{sec:methods}.
Section \ref{sec:numerical} begins by deploying our method on observational data and low dimensional test functions, before tackling our motivating automotive example, a $124$ dimensional problem with $500{,}000$ observations.
%We then empirically illustrate our method's performance on benchmark functions, as well as our motivating automotive example in Section \ref{sec:numerical}
Section \ref{sec:conclusion} concludes the article and overviews promising future work.

\section{Background and Related Work}\label{sec:bg}

We review Gaussian processes before pivoting to gradient sensitivity analysis.

\subsection{Gaussian Processes}\label{sec:bggp}

%A stochastic process with index set $\Xset$ is a rule for assigning to one or more elements of $\Xset$ a joint distribution in a consistent manner. When these joint distributions are Gaussian, the rule must simply specify the mean vector and covariance matrix for any possible combination of inputs. 
%In this article, we will restrict our attention to the case of $\Xset = [0,1]^\nvar$, which accommodates simulators represented as functions of $\nvar$ real-valued input parameters that are independently varied within certain bounds.
%%\footnote{A parameter with any bounds $[a,b]$ can be made to live in $[0,1]$ via the affine transformation $x\to \frac{x-a}{b-a}$.}
%In this case, the mean function is typically constant, $\Esp{}{y(\X)}= \beta_0$, where $y(\X)$ gives the random variables associated with the rows of $\X$, and $\beta_0$ are unknown parameters. 

% Flexible interpolation of computer models is generally the task of a Gaussian process. 
Rather than specifying a functional form, a GP simply defines covariances between input points via some function of the distance between them.
For example: %such as the Gaussian kernel function: 
\begin{equation}
\Var{}{y(\xvec_i), y(\xvec_j)} = \sigma^2 \exp\left\{\frac{-||\xvec_i-\xvec_j||_2^2}{2 l}\right\}, 
\label{eq:iso}
\end{equation}
where the length-scale parameter $l$ controls how quickly correlation decays as the distance between the inputs increases, and the covariance parameter $\sigma$ scales the correlation to turn it into a covariance. 
Broadly speaking, GP kernels differ firstly in how they calculate distance, and secondly in how that distance is translated into a covariance. 
Isotropic kernels such as (\ref{eq:iso}) are those for which every input dimension is treated identically in terms of distance calculations, whereas anisotropic kernels are free to violate this. 
For instance, a tensor-product kernel assigns a different length-scale to each dimension, allowing for correlation to decay at different rates as different parameters are varied. 
Mathematically, this may be expressed as 
\begin{equation}
k(\xvec_i, \xvec_j) := \Var{}{y(\xvec_i), y(\xvec_j)} = \sigma^2 \exp\left\{-\sum_{k=1}^{\nvar} \frac{(x_{i,k} - x_{j,k})^2}{2 l_k}\right\},
\label{eq:sep}
\end{equation}
and evaluation of this kernel between all pairs is usually stored in a \textit{kernel matrix} $\K$.
% as in the case of the so-called separable Gaussian kernel.
Notice that each summand in (\ref{eq:sep}) has a different length-scale $l_k$ in the denominator. Since as $l_k\to\infty$ the contribution of that dimension to the covariance matrix shrinks to zero, the ARD principle \citep[][Ch.~5.1]{Neal1996,Rasmussen2006} argues that dimensions with large length-scales can be ignored. However, technically speaking, there is no guarantee that variable importance decreases monotonically with respect to its length-scale, see \cite[Section 4.1]{LiHsiang2020} and \cite[Section 3.2]{sasl} for counterexamples. Operating somewhat along this principle, \cite{sun2019emulating} and \cite{Katzfuss2020} scale input dimensions according to the inverse of their length-scale before fitting models which involve finding local neighborhoods. This approach will form one of our baselines in Section \ref{sec:methtrans}.

Inference in a GP is typically conducted in a Bayesian manner. Training data, comprising observations $y(\X)$ are collected at certain input locations $\X$ and conditioned on, yielding a posterior GP with modified mean and covariance functions. These latter apply at any desired point $\xvec_{n+1}$ through textbook multivariate Gaussian conditioning:
\begin{align}
    y(\xnew)|\mathbf{y}(\X) &\sim N(\mu_{n+1}, \Sigma_{n+1})& 
    \mu_{n+1} &= \beta_0 + k(\xnew, \X) k(\X,\X)^{-1}(\y - \beta_0\mathbf{1}) \label{eq:gp_mvn} \\
    && \Sigma_{n+1} &= k(\xnew,\xnew) - k(\xnew, \X) k(\X,\X)^{-1} k(\X, \xnew) \nonumber.
\end{align}
The most straightforward way to obtain these quantities involves calculating the Cholesky decomposition of the kernel matrix $k(\X,\X)$, an operation which scales cubically with the number of training locations, $n$, and is computationally intractable when $n$ is in the low thousands.  Much recent work seeks to circumvent this bottleneck.

\subsubsection{Scaling Gaussian Processes to Many Observations}\label{sec:bgbign}

%Usually, an input point will only have high correlation with its neighbors, and negligible correlation with the majority of other points much farther away. However, this correlation will never quite fall to zero. Covariance tapering \citep{Furrer2006} involves smoothly nudging negligible correlations to zero,  accommodating efficient sparse linear algebra. 

Exploiting the fact that an input point will generally only have high correlation with its neighbors, Local Approximate Gaussian Processes \citep[laGP;][Ch.~9.3]{Gramacy2015lagp,Gramacy2020surrogates}, involve constructing a small model at prediction time, incorporating only training points near where a prediction is desired. These points may be selected via Nearest Neighbors (NN) or more sophisticated criteria. The Vecchia approximation \citep{Vecchia1992} also exploits neighborhood structure, but this is used to build a partitioned likelihood.
%, in effect asserting that only nearby points influence one another.
Originally introduced for geospatial data, the Vecchia approximation is most comfortable in low dimensional input spaces, which has motivated a thread of research to adapt it to higher dimensional problems such as surrogate modeling \citep{Katzfuss2020}. That these models select a neighborhood set on the basis of inter-point distances means that proper prewarping could not only give the model a better perspective of distances within the set of local points itself, but also lead to a better set of local points. 
%We use laGP and the Vecchia approximation in our numerical experiments in Section \ref{sec:numerical}. 

%Mercer's Theorem \citep[Ch.~4.3]{Rasmussen2006} tells us that kernel functions are associated with a mapping $\psi$ from $\Xset$ to a new space $\mathcal{U}$, in such a way that the inner product of the new features equals the Kernel matrix, $K_{i,j} = \langle \psi(\xvec_i), \psi(\xvec_j) \rangle_{\mathcal{U}}$. For most popular kernels, this space is infinite dimensional. However, by explicitly choosing the space $\mathcal{U}$ to be finite dimensional but sufficiently rich, the kernel matrix $K$ has a rank bounded by the dimension of $\mathcal{U}$, and can be decomposed efficiently using Woodbury-style identities (and is invertible after adding the diagonal noise matrix). This is the thrust of Fixed Rank Kriging \citep{Cressie2008}. Inducing Point Methods \citep[][Ch.~8]{Smola2001,Snelson2006,Rasmussen2006} are another way of exploiting low rank structure. Instead of calculating the kernel applied to all $\mathcal{O}(n^2)$ pairs of points, the inner product is calculated between all training points and some smaller set of reference locations, knots, or so-called Inducing Points.
Another class of approaches involves choosing a kernel which represents the inner product of a finite-dimensional yet sufficiently rich feature space. Then, the kernel matrix $K$ has a rank bounded by the dimension of the feature space, and can be decomposed efficiently using Woodbury identities. This is the thrust of Fixed Rank Kriging \citep{Cressie2008}. Or, instead of calculating the kernel on all $\mathcal{O}(n^2)$ training pairs, the inner product may be calculated through a smaller set of reference locations, knots, or so-called Inducing Points \citep[][Ch.~8]{Smola2001,Snelson2006,Rasmussen2006}.
%\footnote{Here we characterized these methods as distinct, but Inducing Point Methods may be seen as Fixed Rank Kriging with basis functions of the form $k(\cdot,\xvec$).} 
%The covariance can then be backed out to give a kernel between all $n$ training points, but it will have rank equal to the number of inducing points. 
%Such low-rank methods are referred to as ``sparse", but in the sense of the covariance matrix's spectrum, rather than entrywise sparsity, as is the case with covariance tapering.

The concern with large datasets may seem somewhat antithetical to the idea that each observation was obtained at great computational cost and should be optimally exploited, but there's no other choice in high dimension. Consequently, the adaptation of kernel-based surrogates to high dimensional problems is an area of active research.

\subsubsection{Scaling Gaussian processes to High Dimension}
%There is nothing inherently problematic about GP modeling in high input dimension from a computational perspective, especially for simple isotropic kernels, because arithmetic operations scale cubically in $n$, not $m$.  However, and especially without isotropy, one usually needs an exceedingly large design (large $n$) to create a rich enough training dataset in order to capture signal at high fidelity in that large input space.  Thus, large input spaces tend to be linked to large training datasets.
GP modeling in high dimension requires large designs to accurately capture signal. % in such an expansive design space. 
However, if we assume that the intrinsic dimension of the function is lower than the nominal input dimension, we may be able to get away with a smaller training dataset if a mapping can be learned into this reduced space.
Consequently, many approaches for deploying GP as surrogates in high input dimension settings involve built-in (usually linear) dimension reduction. Perhaps the most straightforward 
%way of conducting linear dimension reduction 
mechanism
involves random projection, as exemplified by Random Embeddings Bayesian Optimization \citep[REMBO]{wang2016bayesian}, and expanded upon in \cite{binois2015warped}. 

Other options include learning projection matrices before fitting a GP on the reduced space. 
In the special case of a one-dimensional reduced space, Bayesian inference via Markov-Chain Monte Carlo has been proposed to learn the low dimensional subspace for both observational data \citep{Choi2011} as well as for computer emulators \citep{Gramacy2012} via Single-Index Models. % which build on the projection pursuit idea. 
\cite{Djolonga2013} combine finite differencing in random directions with low rank matrix recovery to discover the projection matrix. \cite{Garnett2014} give this approach a Bayesian treatment, even proposing an adaptive sampling algorithm to sequentially select informative design points.
Where finite differencing is appropriate, \cite{Constantine2014} propose to deploy adaptive sampling for selecting the low dimensional projection, and also discuss a heuristic for selecting kernel length-scale parameters on the reduced space.

Instead of defining the GP on a low dimensional space, we could split up the dimensions of the input space and define a model on each one.
For instance, \cite{Durrande2010,Duvenaud2011} propose Additive GPs, where the response is modeled as a sum of stochastic processes defined individually for each main effect. The sum can be expanded to include stochastic processes of any interaction level, as detailed in \cite{Durrande2013}, or scalar transformations of the response, as in \cite{LiHsiang2020}.
\cite{delbridge2019randomly} lies at the intersection of random projection and additive kernels: several random projections are combined additively. 
% Further, a kernel is proposed which averages over the random projections. 
%
%Thus far, we have only discussed linear projections. 

%This general idea of regression using sums of low dimensional smoothers dates back to at least Projection Pursuit regression \cite{Friedman1981,Kruskal69}. 

%\subsection{Linear Sensitivity Analysis}

%Axis-Aligned: \cite{sobol2001} is variance based.  

%Linear: Sliced Inverse Regression \cite{Li1991} and Sliced Average Variance Estimation \cite{Cook1991} involves tracing out the trajectory of the input vector $\xvec$ as the \textit{output} $y$ is varied (hence \textit{Inverse}), and then conducting PCA on this trajectory after adjusting for existing correlation in the input to determine the reduced subspace. 

\subsection{Gradient-Based Sensitivity Analysis}
\label{sec:gbsa}

%Several different approaches have been developed to formalize the notion of low intrinsic dimension of a black-box function $f: \mathbb{R}^m \rightarrow \mathbb{R}$, whose evaluations may require running an expensive computer simulation. This article is most concerned with methods that analyze the gradient of the target function, but it should be noted that other interesting approaches have been suggested, such as those based on characterizing the distribution of response given the important low dimensional subspace \citep{Li1991,Cook1991} or based on variance \citep{sobol2001}. The relationship between gradient-based methods and variance-based methods is explored by \cite{Kucherenko2016}. 

%If derivatives of the simulator are available with respect to input parameters, a natural way to define importance of the inputs is via the magnitude of $\frac{\partial f(\xvec)}{\partial x_i}$ since this quantity tells us how much the output changes as the input variable $i$ is perturbed, assuming the input scales are comparable. However, gradients are a fundamentally local quantity, so are on their own ill-defined as a global sensitivity metric. Global analysis results from defining some manner of aggregating the function's gradients throughout the input space. The techniques we review below, as well as those we deploy in our methodology, are defined as expectations of gradients over the input space, but it is worth noting that applications may motivate other approaches, for instance the maximum derivative $\max_{\xvec \in \Xset}|\frac{\partial f(\xvec)}{\partial x_i}|$. 
If derivatives of the simulator are available with respect to input parameters, a natural way to define importance of the inputs is via the magnitude of $\frac{\partial f(\xvec)}{\partial x_i}$ since this quantity tells us how much the output changes as input variable $i$ is perturbed, assuming the input scales are comparable. Global sensitivity analysis proceeds by defining some method of aggregating such averaging as proposed by \cite{sobol1995}, who used $\mathbb{E}\{(\frac{\partial f(\xvec)}{\partial x_i})^2\}$, estimated via Finite Differencing, as a measure of variable importance for screening purposes. \cite{DeLozzo2016} describe a GP based estimator for this quantity. But we are interested in \textit{directions} of importance, which may be defined by those with large average \textit{directional} derivatives.

%Computing this estimate involves finite differencing, which may be infeasible for stochastic or numerically sensitive functions.
%To this end, \cite{DeLozzo2016} proposed GP-based estimates for derivatives.
%Since these screening methods retain only inputs which have large derivatives, a natural extension is to consider \textit{directions} within the input space which have large \textit{directional} derivatives, effectively relaxing the constraint that these directions be aligned with the input axes. In this case, we are looking for linear combinations of input variables along which the function changes most drastically. 

\begin{figure}
    \centering
    \includegraphics[scale=0.4,trim=15 40 0 15]{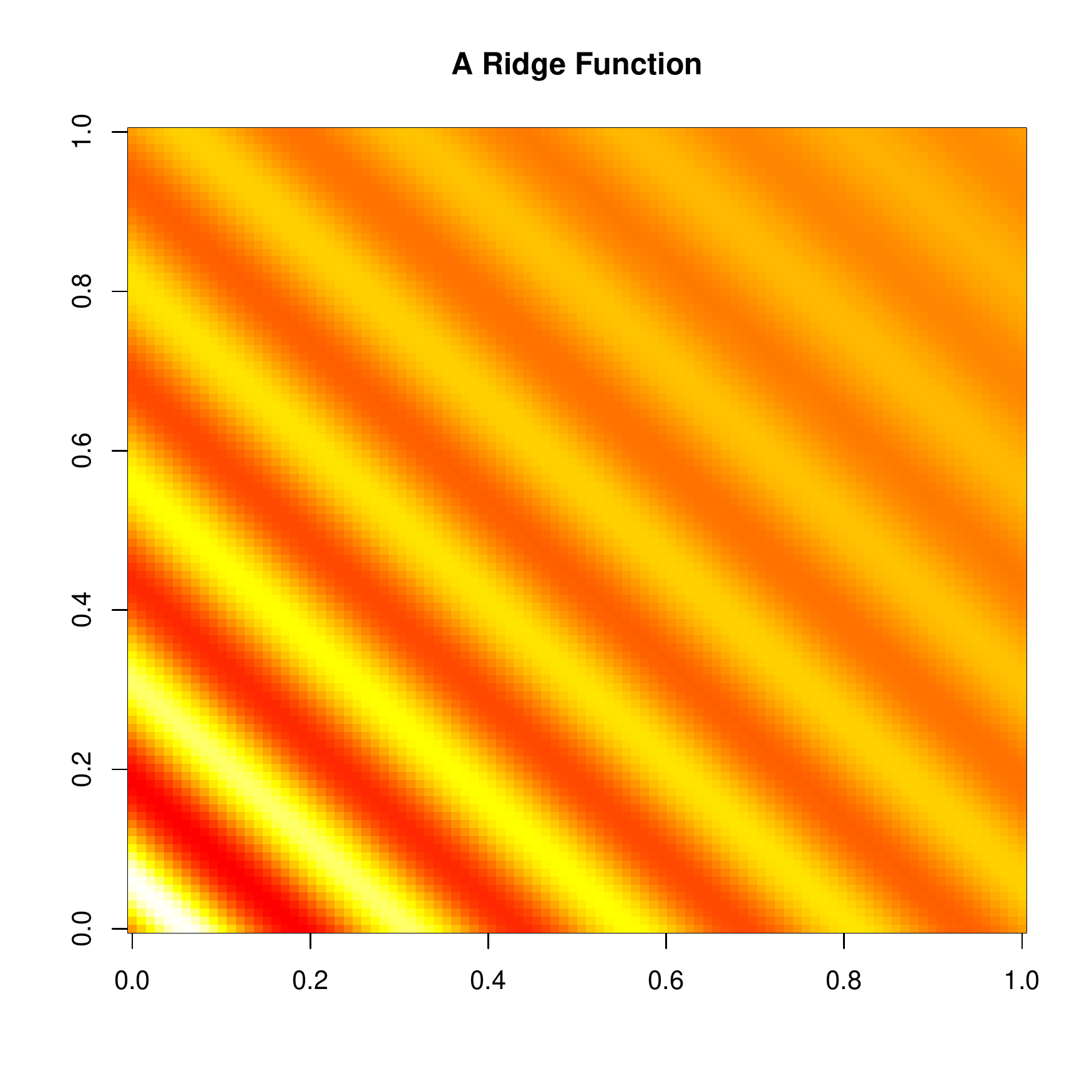}
    \includegraphics[scale=0.4,trim=10 40 0 15]{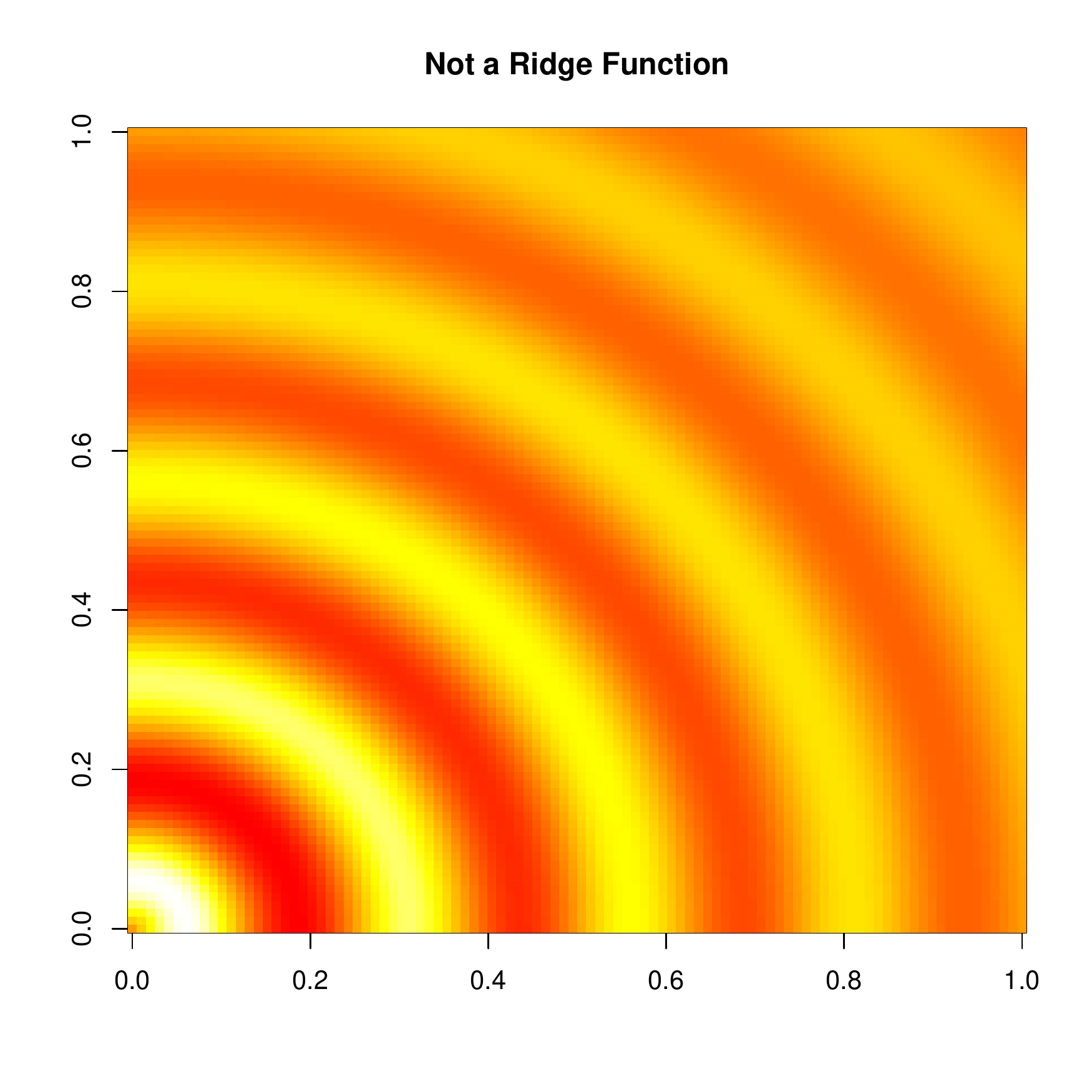}
    \caption{Heat plots of {\em left:} $f(z) = \sin(z)\cos(z)e^{-\frac{z}{10}}$, with $z=x+y$; and {\em right:} $z=\sqrt{x^2+y^2}$.}
    \label{fig:ridge}
\end{figure}

Functions varying \textit{only} in certain directions are called Ridge Functions, and thus have the form $f(\xvec) = g(\mathbf{A} \xvec)$, where $\mathbf{A} \in \mathbb{R}^{\nred\times \nvar}$, $g$ is any function on $\mathbb{R}^\nred$, and $\nred < \nvar$.  As a modeling device,
ridge functions have inspired a number of nonlinear
statistical predictors, including projection pursuit \citep{Friedman1981}. In the ridge function framework, dimension reduction is assumed to be linear, but the actual function on the low dimensional space need not be. The left panel of Figure \ref{fig:ridge} shows the ridge function $f(x) = \sin(x+y)\cos(x+y)e^{-\frac{x+y}{10}}$. Eponymous ridges are visible as constant diagonal bands in the heat plot.  Here, $\mathbf{A} = [1 \hspace{0.5em} 1]$, and $g(z) = \sin(z)\cos(z)e^{-\frac{z}{10}}$. Note, however, that ridge functions cannot exhibit ``curvy" ridges, as in the right panel. From the ridge function perspective, the right image represents a two dimensional function, even if it depends only the one dimensional quantity $\sqrt{x^2+y^2}$.

The Active Subspace method \citep[AS;][]{Constantine2015} provides a way to view functions as being ``almost" ridge functions. This analysis considers the expected gradient outer product matrix with respect to some measure $\nu$:%
\begin{equation}\label{eq:C}
    \C = \Esp{\nu}{\nabla f\nabla f^\top} = \int \nabla f\nabla f^\top\ d \nu\,.
\end{equation}
Functions are said to have an AS when they change \textit{mostly} rather than uniquely along a small set of directions, formalized in the sense that $\C$ has a gap in its eigenvalues.
The eigenspace associated with the eigenvalues that make the cut are those directions in which large gradients are ``often" pointed, relative to the measure $\nu$. 

In this article, the measure with respect to which the AS is defined will either be the Lebesgue measure $\nu_l$ or the sample probability measure $\nu_s$, which is given by $\nu_s(\mathcal{A}) = \frac{1}{n}$ if $\mathcal{A} = \{\xvec_i\}$ for any sample point $\xvec_i$ (such that taking the expectation of some quantity with respect to this measure is simply the average of that quantity observed at the sampling locations). We use $\nu$ to denote a generic probability measure.

Readers familiar with techniques such as PCA that analyze the spectrum of the \textit{covariance} matrix might expect us instead to be interested in 
\begin{equation*}
    \Esp{\nu}{(\nabla f - \Esp{\nu}{\nabla f})(\nabla f - \Esp{\nu}{\nabla f})^\top} =
    \Esp{\nu}{\nabla f\nabla f^\top} -
    \Esp{\nu}{\nabla f}\Esp{\nu}{\nabla f}^\top,
\end{equation*}
the only difference being that the mean gradient is subtracted prior to the outer product. 
%(and since the modification is of rank 1, the action of this quantity and the active subspace matrix differ only in one direction). 
However, in the case of analyzing gradients rather than data points, the average gradient contains useful information about the function. This is to the extent that \cite{Lee2019} even proposes adding the $\Esp{\nu}{\nabla f}\Esp{\nu}{\nabla f}^\top$ term above rather than subtracting it to enhance the influence of that direction. 

Analytically computing the integral defining $\C$ is not possible for a general blackbox $f$. However, if the gradient may be evaluated at arbitrary input locations, a Monte Carlo estimator may be formed by first sampling $B$ many vectors $\xvec_i \sim \nu$, and then computing $\frac{1}{B} \sum_{i \in \{1,\ldots, B\}} (\nabla f)(\xvec_i)(\nabla f)(\xvec_i)^\top$. As with the axis-aligned sensitivities, we can of course use finite-difference approximations; \cite{Constantine2015} analyzes the effect of numerical error in this step on the quality of the overall estimate of $\C$. 

 %Zhou, et al.~\cite{Zhou2013}
 
In situations where finite differencing is not appropriate, the derivative may again be estimated via nonparametric methods \citep{Othmer2016,palar2017,Palar2018}. Given a GP posterior with constant prior mean $\beta_0$ on $f$, a natural way to estimate $\C$ is to use the posterior mean of the integral quantity it is defined by (Eq.~\ref{eq:C}), which is now a random variable as we are conducting Bayesian inference. Assuming a sufficiently smooth kernel function, the gradient vector at any point $\mathbf{x}^*$ has a multivariate Gaussian posterior $\nabla f(\mathbf{x}^*) \sim N( \mu_{\nabla},  \Sigma_{\nabla})$, where
	\begin{align*}
		\mu_\nabla &= \mathbf{K}_{[\nabla, X]} \mathbf{K}_{[X,X]}^{-1} (\mathbf{y}-\beta_0)\,, \\
		\mbox{and } \quad \Sigma_{\nabla} &= K_{[\nabla,\nabla]} - K_{[\nabla, X]}\mathbf{K}_{[X,X]}^{-1}K_{[X, \nabla]}\, .
	\end{align*}
Above, $\mathbf{K}_{[\nabla, X]}$ represents the cross-covariance matrix between the gradient at $\mathbf{x}^*$ and the observed outputs $\mathbf{y}$, $\K_{[X,X]}$ that between the outputs $\mathbf{y}$ at each training location, and $\mathbf{K}_{[\nabla, \nabla]}$ represents the prior covariance matrix of the gradient vector. These quantities are easily derived in terms of derivatives of the kernel function $k$ \citep[Ch.~9]{Rasmussen2006}, and was used as early as \cite{Morris1993} to exploit observed derivative information to improve a computer experiment response surface. We will use these facts to simplify the desired expectation:
\begin{align*}
    \Esp{f}{\C_{\nu}|\mathbf{y}} &= \Esp{f}{\Esp{\xvec\sim\nu}{\nabla f(\xvec)\nabla f(\xvec)^\top}|\mathbf{y}} \\
    & = \Esp{\xvec\sim\nu}{\Esp{f}{\nabla f(\xvec)\nabla f(\xvec)^\top|\mathbf{y}}} 
    = \Esp{\xvec\sim\nu}{\Sigma_{\nabla}(\xvec) + \mu_\nabla(\xvec) \mu_\nabla(\xvec)^\top}\,.
\end{align*}
For general $\nu$, this expression may be evaluated via Monte Carlo. \cite{sasl} provided closed forms for when $\nu$ is the Lebesgue measure on $[0,1]^\nvar$ (denoted $\nu_l$) and $k$ is Gaussian (\ref{eq:iso}--\ref{eq:sep}) or Mat\'ern with smoothness $\frac{3}{2}$ or $\frac{5}{2}$. The quantities above depend on the choice of kernel  hyperparameters, which must be estimated. We prefer  maximizing the marginal likelihood, but other options work \citep{Fukumizu2014}.

The quantity $\C$ was studied for observational data as early as \cite{samarov1993}. 
Kernel based estimates were proposed by \cite{Fukumizu2014} with respect to the sample measure $\nu_s$, and deployed by   
\cite{Liu2017} to reduce the dimension of a tsunami simulator.  Authors have also considered second order derivatives. \cite{li1992} proposes looking at Hessian eigen-decompositions in Principal Hessian Directions as well as a method to estimate the Hessian itself using Stein's Lemma, effectively calculating the cross-covariance between the response and the outer product of the input vector. 
%
%Because they take the kernel regression rather than Gaussian process perspective, they do not have a marginal. likelihood to choose hyperparameters. Consequently, they use an isotropic kernel, and choose its length-scale using CV. 
%
For more on GSA, see \cite{Iooss2015}.

\section{Methodology}  \label{sec:methods}

We first discuss how to turn a sensitivity analysis into an input warping before discussing how to fit local models in the warped space.

\subsection{Warping}
%\subsection{Active Subspace after Transformation}
\label{sec:methtrans}

Here we propose the heuristic of using the warping such that running the sensitivity analysis again afterwards would result in all directions being equally important.
In the case of ARD, this would amount to conducting a warping such that the optimal length-scales are all equal to 1, while in the case of AS, $\C = \mathbf{I}$.
In both of these cases the transformation is linear, and thus can be represented by a matrix $\LL$.
%, though one could certainly consider nonlinear sensitivity analysis methods for which this would not be the case.
The matrix 
$\LL$  should premultiply each design point $\mathbf{z}_i=\LL\xvec_i$, which looks like $\Xtrans=\X\mathbf{L}^\top$ when the design points are stacked in the canonical design matrix $\X \in \mathbb{R}^{n\times\nvar}$.
This process may be seen as decomposing the black-box $f$ into two parts: a linear transformation $\mathbf{L}$ and a nonlinear function $g$.
Here, $g$ is the function upon which we are actually doing regression when we fit $\mathbf{y}$ to $\Xtrans$.
%We next describe how $\LL$ can be formed via ARD or AS sensitivity.

\subsubsection{Bandwidth and Range Scaling}

When using the separable Gaussian kernel (Eq.~\ref{eq:sep}),  a length-scale of $l_k$ for input variable $k$ living in $[0,1]$ is equivalent to using a length-scale of $l_k = 1$ and a domain of $\left[0,\frac{1}{\sqrt{l_k}}\right]$. Therefore, scaling each input dimension by the root of its estimated length-scale would achieve our desired result. This is because fitting a GP to the scaled input-output relationship would result in length-scale estimates equal to 1. %for all input variables. % (modulo numerical issues).

\begin{algorithm}[ht!]
	\caption{Bandwidth Scaling} 
	\hspace*{\algorithmicindent} \textbf{Given:} Data $\X, \y$, Bags $B$, Bag size \texttt{nsub}, Sample Size $\nsamp$,%\emph{!$N$?!}
	\begin{algorithmic}[1]
	    \For {$b \in \{1, \ldots, B\}$}
	        \State $\mathcal{I} \sim \textrm{Cat}\{1, \ldots, N\}$\Comment{Subsampling}
	        %\State $\mathcal{I} \sim \textrm{Sample}(\nsamp, \texttt{nsub})$\Comment{Subsampling without replacement.}
	        \State $\hat{\boldsymbol\theta}_{b} \gets \underset{\theta}{\textrm{argmin}}\, \mathcal{L}_{GP}(\mathbf{y}_{\mathcal{I}}|\theta)$\Comment{Optimize GP Likelihood wrt $\boldsymbol\theta$}
		\EndFor
		\State $\hat{\boldsymbol\theta} \gets \frac{1}{B} \sum_{\mathcal{B}} \hat{\boldsymbol\theta}_{\mathcal{B}}$
		\State $\LL \gets \textrm{diag}(\hat{\boldsymbol\theta})$\Comment{Place Estimates in a Diagonal Matrix}
		\State $\Xtrans \gets \X \LL^\top$
	\end{algorithmic} \label{alg:bandscale}
\end{algorithm}

%Since ARD is an axis aligned method, $\LL$ will be a diagonal matrix, resulting in a scaling of the input space along each dimension, and will simply have the square root of the length-scales on its diagonal:
Since we are just scaling the input space, $\LL$ will be a diagonal matrix with nonzero elements given by the inverse root of the length-scales:
$\LL_{\mathrm{ARD}} = \mathrm{Diag}\left(\frac{1}{\sqrt{l_1}}, \frac{1}{\sqrt{l_2}}, \cdots, \frac{1}{\sqrt{l_\nvar}}\right)$.
% \begin{equation}
%     \LL_{\mathrm{ARD}} =
%     \begin{bmatrix}
%     \cccccccccccc & 0 & \ldots & 0\\
%     0 & \frac{1}{\sqrt{l_2}} & \hdots &0\\
%     \vdots & \vdots & \ddots & 0\\
%     0 & 0 & \ldots & \frac{1}{\sqrt{l_{\nvar}}}
%     \end{bmatrix} 
% \end{equation}
In \cite{Gramacy2020surrogates} and \citet{Cole2020} this is treated as a preprocessing step, performed once before deployment within local models, while in \cite{Katzfuss2020} this scaling is iteratively updated as the marginal likelihood is optimized and length-scale estimates change. \citeauthor{Cole2020} attributed the idea to Derek Bingham, who called it ``stretching and compressing". 

Other approaches of input variable sensitivity could be considered in developing transformations. As recommended by an anonymous reviewer, we will consider another measure of sensitivity to be the range of the GP posterior surface fit to data projected onto a given axis. In particular, to determine the \textit{range sensitivity} of variable $i$, we first fit a one dimensional GP regression on $\mathbf{X}_i$ vs $\mathbf{y}$. Then, the sensitivity is defined as the range of the posterior surface of that GP, that is to say, as $\underset{x_1,x_2 \in [0,1]}{\max} |\hat{f}(x_1) - \hat{f}(x_2)|$ where $\hat{f}$ is the posterior predictive mean. This is a nonconvex optimization problem which we solve approximately by initializing $x_1$ and $x_2$ to be the i'th coordinates of those design points corresponding to the largest and smallest observed $y$ values and then applying a quasi-Newton method (L-BFGS-B) refinement.
%To incorporate truncation into this framework, we suggest dropping the $\nvar - \nred$ variables with the largest length-scales, resulting in a still diagonal but now rectangular $\LL$ retaining only the pertinent rows. This procedure is given in Algorithm \ref{alg:bandscale}, where selection of $\nred$ is achieved via $k$-Nearest-Neighbors, as detailed in Appendix \ref{sec:aploc}.

%A simple alternative to a full rotation is a rescaling of each input dimension, which may be viewed as a diagonal sensitivity matrix. {\em (Nate, have we talked about using the C matrix to rotate and rescale yet?  Maybe we need our own section here for this or maybe this goes after 3.2?  On the other hand, perhaps a discussion of what other have done in the past belongs in Section 2, and what we do here -- after talking about rescaling via C -- is relate back to that discussion?)}. Previously \cite{Gramacy2020surrogates,Katzfuss2020} authors use GP estimates of separable length-scales (\ref{eq:sep}) to determine the scaling, setting each {\em (Nate: what?)} to the inverse of {\em (Nate: square root?)} the estimated length-scale associated with that input dimension. 

\subsubsection{Active Subspace Rotation}

%In this article, we propose to perform the linear transformation on the input space given by the matrix $\mathbf{L} = \Lambda_{r}^{1/2} \mathbf{U}_r^\top$ where $\mathbf{U}_r\in\mathbb{R}^{\nvar\times r}$ is the matrix with columns giving the eigenvectors of $\C$ associated with its $r$ largest eigenvalues and $\Lambda_{r}^{1/2}$, a diagonal matrix containing the square root of those same eigenvalues. In terms of a design matrix $\X$ with rows given by $\mathbf{x}_i\ \in\mathbb{R}^\nvar$, this amounts to computing the eigendecomposition $\C = \mathbf{U}\Lambda\mathbf{U}^\top$, and then post-multiplying our design matrix by the factor $\Xtrans = \X\mathbf{L}^\top$.  {\em (Nate, maybe we need a more deliberate, concrete description here, either mathematically or in code. In a way, this looks like what is in 3.4 as pseudocode.) }

%We show below that if our $\C$ estimate were exact, this would lead to the local Gaussian processes being fit on a function which has as its active subspace matrix the identity.\footnote{Assuming that $r \leq \textrm{rank}(\C)$.}  

\begin{figure}[ht!]
    \centering
    \vspace{0.25cm}
    \includegraphics[scale=0.26]{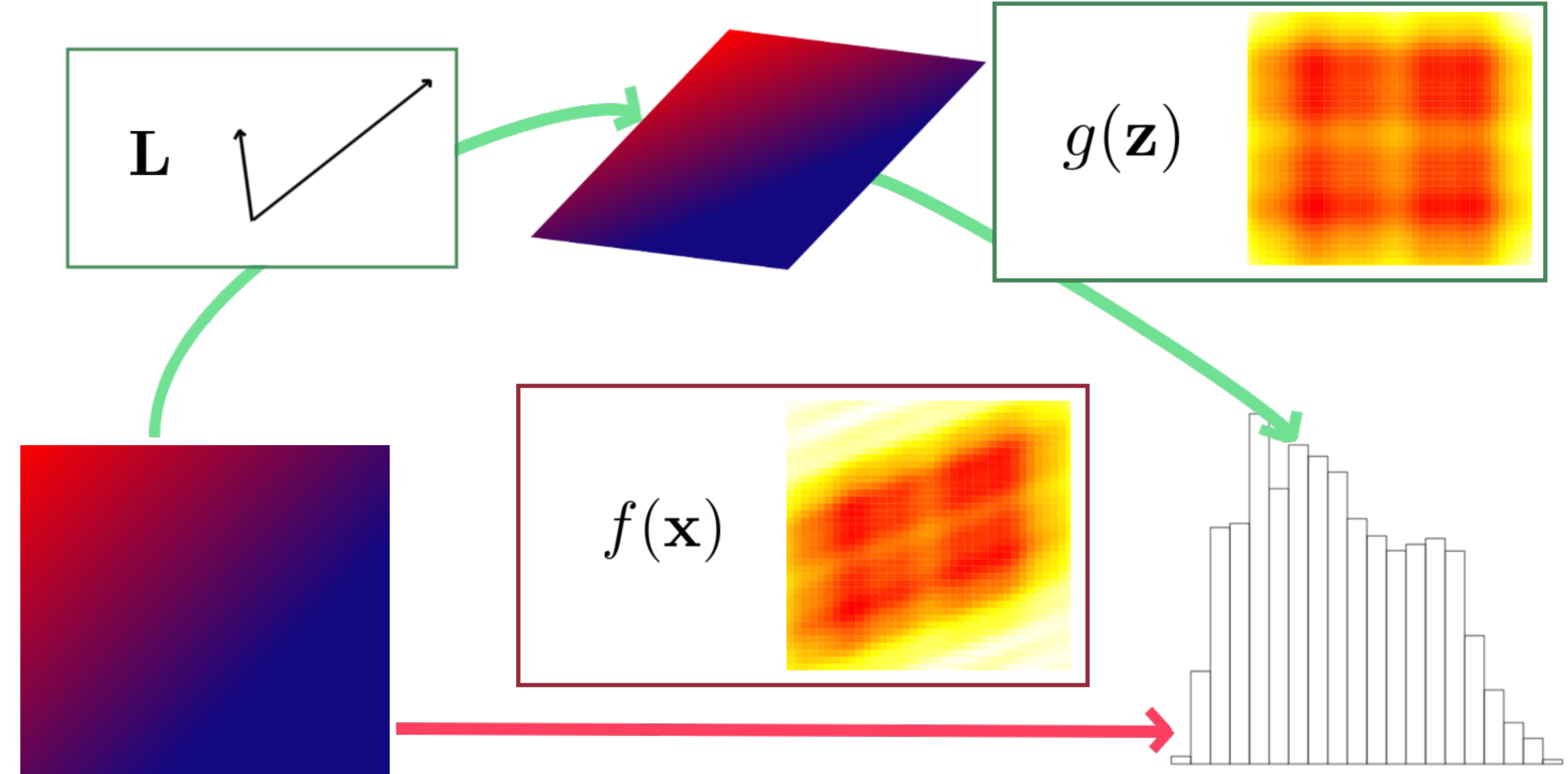}
    \vspace{0.1cm}
    \caption{The function $f$ (bottom, red line) with a nontrivial AS maps from $[0,1]^2$ to $\mathbb{R}$. It may alternatively be viewed as a linear scaling $\mathbf{L}:[0,1]^2\to\mathbb{R}^2$, followed by a function $g$ with all directions of equal importance (top, green lines). Before preprocessing, regression is on $f$; afterwards on $g$.}
    \label{fig:decomp}
\end{figure}

In the case of a known AS matrix $\C$, the transformation $\LL$ which satisfies our desire to ``undo" the sensitivity analysis is given by $\mathbf{L} = \Lambda^{1/2} \mathbf{U}^\top$, where $\mathbf{U}\in\mathbb{R}^{\nvar\times \nvar}$ is the matrix with columns giving the eigenvectors of $\C$ and $\Lambda^{1/2}$ a diagonal matrix containing the square root of the eigenvalues.
To how that this warping satisfies our heuristic, recall that $f(\xvec) = g(\mathbf{L}\xvec)$, and let  $\nu_{\mathbf{z}}$ be the measure implied on $\mathbf{z}:=\mathbf{L}\xvec$ by $\nu$.
\begin{align}
		& \Esp{\nu}{ \nabla_{x} f(\xvec) \nabla_{x} f(\xvec)^\top} = \Esp{\nu}{ \nabla_{x} g(\mathbf{L} \xvec) \nabla_{x} g(\mathbf{L} \xvec)^\top} \nonumber \\
		& \iff  \Esp{\nu}{ \nabla_{x} f(\xvec) \nabla_{x} f(\xvec)^\top} = \Esp{\nu}{ \mathbf{L}^\top(\nabla_{\mathbf{L}x} g(\mathbf{L}\xvec)) (\nabla_{\mathbf{L}x} g( \mathbf{L}\xvec))^\top\mathbf{L}} \nonumber \\
		& \iff \Esp{\nu}{ \nabla_{x} f(\xvec) \nabla_{x} f(\xvec)^\top} =  \mathbf{L}^\top\Esp{\nu}{\nabla_{\mathbf{L}x} g(\mathbf{L}\xvec) \nabla_{\mathbf{L}x} g( \mathbf{L}\xvec)^\top} \mathbf{L} \nonumber \\
		& \iff \mathbf{U}\Lambda\mathbf{U}^\top =  \mathbf{U}\Lambda^{\frac{2}{2}}\Esp{\nu}{\nabla_{\mathbf{L}x} g(\mathbf{L}\xvec) \nabla_{\mathbf{L}x} g( \mathbf{L}\xvec)^\top} \Lambda^{\frac{1}{2}}\mathbf{U}^\top \nonumber\\
		%& \iff \Lambda =  \Lambda^{\frac{1}{2}}\Esp{\nu}{\nabla_{\mathbf{L}x} g(\mathbf{L}\xvec) \nabla_{\mathbf{L}x} g( \mathbf{L}\xvec)^\top} \Lambda^{\frac{1}{2}} \\
		&\iff \textbf{I} = \Esp{\nu}{\nabla_{\mathbf{L}x} g(\mathbf{L}\xvec) \nabla_{\mathbf{L}x} g( \mathbf{L}\xvec)^\top}, \nonumber
\end{align}
% or alternatively
%\begin{equation}
%    \Esp{\nu_{\mathbf{z}}}{\nabla_{\mathbf{z}} g(\mathbf{z}) \nabla_{\mathbf{z}} g( \mathbf{z})^\top} = \mathbf{I}\,,
%\end{equation} 
% \emph{!ance?!}
or alternatively $\Esp{\nu_{\mathbf{z}}}{\nabla_{\mathbf{z}} g(\mathbf{z}) \nabla_{\mathbf{z}} g( \mathbf{z})^\top} = \mathbf{I}$.
Consequently, all directions are of equal importance globally, and the local model is freed to concentrate on local information. The decomposition is illustrated in Figure \ref{fig:decomp}, which shows the trajectory from simulator input to simulator output in two different ways. 
The bottom of the figure shows the standard modeling approach, where the black-box simulator maps directly from the input space to the scalar response in an anisotropic manner. The top shows our proposed decomposition, where first a linear transformation maps the input hypercube into a polytope defined by the sensitivity analysis, and second the now isotropic nonlinear function may be modeled by local predictors. This procedure is delineated in Algorithm \ref{alg:asrot}, which defines a family of warpings parameterized by the measure $\nu$. In this article, we will study the transformations $\LL_{l}$, associated with the Lebesgue measure, and $\LL_{s}$, associated with the sample measure.

\begin{algorithm}[ht!]
	\caption{Active Subspace Rotation} 
	%\hspace*{\algorithmicindent} \textbf{Given:} Data $\X, \y$, $\nu \in \{\textrm{Lebesgue}, \textrm{Sample}\}$, Bags $B$, Bag size \texttt{nsub}, Sample Size $\nsamp$ %\emph{!$N$?!}
	\textbf{Given:} Data $\X, \y$, $\nu \in \{\textrm{Lebesgue}, \textrm{Sample}\}$, Bags $B$, Bag size \texttt{nsub}, Sample Size $\nsamp$ %\emph{!$N$?!}
	\begin{algorithmic}[1]
	    \For {$b \in \{1, \ldots, B\}$}\Comment{Subbagging Iteration}
	        \State $\mathcal{B} \sim \textrm{Cat}(\{1, \ldots, N\},\texttt{nsub})$
	        \Comment{Subsampling}
	        \State $\hat{\boldsymbol\theta}_{\mathcal{B}} \gets \underset{\theta}{\textrm{argmin}}\, \mathcal{L}_{GP}(\mathbf{y}_{\mathcal{B}},\mathbf{X}_{\mathcal{B}}|\theta)$
	        \Comment{Optimize GP Likelihood wrt $\boldsymbol\theta$}
	        \State $\hat{\C}_{\mathcal{B}} \gets \Esp{\nu}{\nabla f(\xvec)\nabla f(\xvec)^\top|\mathbf{y}_{\mathcal{B}}}$
	        \Comment{Subset estimate of $\C$}
		\EndFor
		\State $\hat{\C} \gets \frac{1}{B} \sum_{\mathcal{B}} \hat{\C}_{\mathcal{B}} $
		\State $\mathbf{U},\boldsymbol{\Lambda} \gets \texttt{eigendecomp}(\hat{\C})$
		\State $\LL \gets \boldsymbol{\Lambda}^{\frac{1}{2}}\mathbf{U}^\top$
		\State $\Xtrans \gets \X \LL^\top$
	\end{algorithmic} \label{alg:asrot}
\end{algorithm}

\subsubsection{Truncation}

%To incorporate truncation into this framework, we suggest dropping the $\nvar - \nred$ variables with the largest length-scales, resulting in a still diagonal but now rectangular $\LL$ retaining only the pertinent rows. This procedure is given in Algorithm \ref{alg:bandscale}, where selection of $\nred$ is achieved via $k$-Nearest-Neighbors, as detailed in Appendix \ref{sec:aploc}.  
Once a transformation $\LL$ is calculated, we may additionally select a truncation dimension, creating another, more parsimonious class of options for the warping. Determining the appropriate amount of such truncation depends on what local predictor is to be applied downstream, on the warped (and lower dimensional) inputs.  We follow the approach outlined by \cite{Fukumizu2014}, which is actually designed to estimate kernel hyperparameters but it is easily adapted to any low-dimensional parameter, like model complexity.  Our pseudo-code in that setting is provided in Algorithm \ref{alg:trunc}.  Notice that the method involves NN, however this is just one of many possible downstream models, a discussion we shall table for the moment.
We take the same approach to truncation regardless of which GSA method gave rise to $\LL$.
%In particular, candidate truncation dimensions $\nred$ from $5$ to \change{$\texttt{min}(\nvar, 25)$} are evaluated{\em (Nate, I still don't understand this.  MIND and MAXD in the algorithm aren't defined or discussed.  I don't understand the significance of 5 and 25.  5 seems too large for some of our smaller examples, and 25 to small for the larger ones.)}, deploying NN on each before computing the sum of squared residuals.
In particular, NN is applied to each candidate dimension, and the sum of squared residuals computed.
Rather than simply choosing that dimension which minimized error magnitude, we found that optimizing the Bayesian Information Criterion (BIC) was superior. In calculating BIC, we treated the dimension of the NN model as the number of parameters it had and endowed it with a Gaussian error structure. 

\begin{algorithm}[ht!]
	\caption{Dimension Selection} 
	\hspace*{\algorithmicindent} \textbf{Given:} Rotated Design Matrix $\Xtrans$, search interval [\texttt{MIND}, \texttt{MAXD}].
	\begin{algorithmic}[1]
	\For {$\nred^* \in \{\texttt{MIND}, \ldots, \texttt{MAXD}\}$}
	    \State $\Xtrans_{\nred^*} \gets \Xtrans[,1:\nred^*]$
	    \State $\texttt{mse}[\nred^*] \gets \texttt{mean}(\texttt{resid}(\texttt{KNN}(\Xtrans_{\nred^*}, \y))\texttt{\textasciicircum 2})$
	    \Comment{$\kappa$-Nearest Neighbors}
	    \State $ \texttt{bic}[\nred^*] \gets \nsamp \log(\texttt{mse}[\nred^*]) + \nred^* \log(\nsamp) $
	\EndFor
	\State $\nred \gets \underset{\texttt{MIND}\leq \nred^* \leq \texttt{MAXD}}{\textrm{argmin}} \texttt{bic}[\nred^*]$
	\end{algorithmic} \label{alg:trunc}
\end{algorithm}

%Arguably, this approach is somewhat mismatched if the downstream local predictor is not based on NN.
%And, indeed, if the practitioner is entering the modeling process with a particular local model in mind -- our preferred options are enumerated next in Section \ref{sec:methloc} -- it would be perfectly reasonable to select the truncation dimension via CV applied directly to that model.
%However, this would entail additional computational overhead, as NN is about as simple as it gets, and would also mean that the dimension needs to be reselected if a different local modeling approach is desired.
In our experiments (Section \ref{sec:numerical}), all of our local models use the same truncated dimension size $\nred$ selected by Algorithm \ref{alg:trunc}.
Other approaches still are certainly possible.
For instance, \cite{Constantine2015} suggests manual examination of $\C$'s spectrum for a gap, though such human intervention may be at odds with the otherwise hands-off, automated approach implied by the surrogate modeling context.

%It is important to ensure that $\LL$ is properly ordered before we do truncation, to ensure that we are ignoring the least important directions when taking only the first $\nred$ columns of $\Xtrans$. For Bandwidth Scaling, this implies sorting the length-scales in increasing order (or the inverse-roots in decreasing order) along the diagonal of $\LL$, while for AS prewarping, this involves making sure the elements of $\Lambda$ and $\mathbf{U}$ correspond to eigenpairs in decreasing order. This way, we make sure to drop the variables with the $\nvar-\nred$ largest length-scales for $\LL_{\mathrm{\mathrm{ARD}}}$, and those directions associated with the $\nvar-\nred$ smallest eigenvalues for $\LL_{l}$ and $\LL_{s}$.

\subsubsection{Scaling Up}
\label{sec:scale}

GP-based estimates of the active subspace carry the GP's computational burdens, and are limited to comparatively small datasets, just as the GP itself is. We mitigate this via a subbagging approach \citep{Breiman1996,Zhao2018}. Given a subbag size $n_B<\nsamp$ and a number of subbags $B$, we simply sample $n_B$ many datapoints at random from our input-output pairs before fitting a GP and developing an estimate of $\C$ based on those data alone. This is repeated $B$ times, and each estimated $\C_b$ is combined via averaging to form our estimator $\frac{1}{B}\sum_{b=1}^B \C_b$.
Since we are executing the cubic cost GP operations not on $\nsamp$ data but on $n_B$ data, the overall computational expense is significantly less on our applications despite the fact that the procedure must be repeated several times. Furthermore, this is an embarrassingly parallel task. Of course, this comes at the cost of estimation error, and, to our knowledge, the impact of such subsampling on the concentration rate of the estimate of $\mathbf{C}$ is an open question. We find that it works in practice in Section \ref{sec:numerical}.

%Since the downstream local emulation approaches (Section \ref{sec:methloc}) scale to large datasets, we want to ensure that our method of preprocessing is scalable as well. Our GP estimates of the AS matrix and length-scales suffer the bottleneck of decomposing large covariance matrices, so a direct application is not feasible. Instead, we will employ a subbagging technique \citep{Breiman1996,Zhao2018}. Given a number of subbags $B$ and subbag size $n_B \ll n$, we sample $B$ samples of size $n_B$, then calculate each of the $B$ AS matrices $\C_b$ as described in Section \ref{sec:methcompC}, before averaging them (See Algorithms \ref{alg:bandscale}--\ref{alg:asrot}).  In the context of ARD, subsampling can be shown to provide consistent estimates of length-scale relative to MLEs estimated from the full data \citep{liu:2014,liu:hung:2015}.  Extending to AS is natural, but as yet unproven in practice.
%With an estimate of our sensitivity analysis in hand, we can go about exploiting it.

\subsection{Local Modeling}\label{sec:methloc}

For some regression methods, such as the basic linear model, linear transformations such as those we have described in this section so far would have no nontrivial impact. However, this is certainly not the case for local models, which are influenced in two major ways, namely by altering the partitioning scheme and by changing the default distance metric. Before we see exactly how, we provide an overview of the particular local models we prefer; the Supplementary Material provides further detail.

The simplest of these is NN. To predict at $\xnew$, NN determines the $k$ closest training locations to $\xnew$, then averages their responses to obtain a prediction. It is thus affected by the linear warping through a warped definition of ``closest", which thus alters the points which are being averaged for each prediction.

% \emph{!overlap with kernel?}
The laGP method also operates by building a prediction set at $\xnew$. And, just like NN, it begins with some number $\kappa$  of nearest neighbors to $\xnew$. Next, however, points are added to that set based on how useful they will be for prediction as measured by an acquisition criterion built on a GP. This GP is grown until some pre-specified ``max'' size. Both the conditioning set(s) (like NN), and the local kernel function are a influenced by the linear pre-warping. 

The Vecchia approximation is a related but distinct idea. Unlike NN or laGP, which create local models at prediction time, the Vecchia approximation specifies a single generative story for the data. 
Each datapoint, rather than being conditioned upon all other training data, is instead conditioned on a cascade of subsets, assumed conditionally independent of all others.
This requires the data be ordered, making the assumption that any data point is conditionally independent of all those data that come after it in the order.
%given to those points in its conditioning set, which are constrained to come before.
Since vector data in general have no natural ordering, one is generally imposed by sorting along a given axis or finding an ordering that best encodes input distances \citep{Guinness2018}.
%In Euclidean spaces of dimension greater than 1, there is no canonical ordering to turn to, so in practice an order is imposed on the data.
%This can be as simple as choosing to sort along whichever axis is perceived to be of greatest importance.
%In this article, we instead use the maximin ordering proposed by \cite{Guinness2018}, which aims to minimize the discrepancy between  points as inferred by the ordering as measured in Euclidean space.
The Vecchia approximation stands to benefit from an improved ordering (and kernel structure) via prewarping.  %]on top of neighborhood selection and distance metric change when prewarping is applied to it.

\subsubsection*{Illustrating Influence on Neighborhood Selection}

%\emph{!replace most nearest neighbors with NN?!}

\begin{figure}
    \centering
    \vspace{-0.25cm}
    \raisebox{-0.5\height}{\includegraphics[scale=0.35]{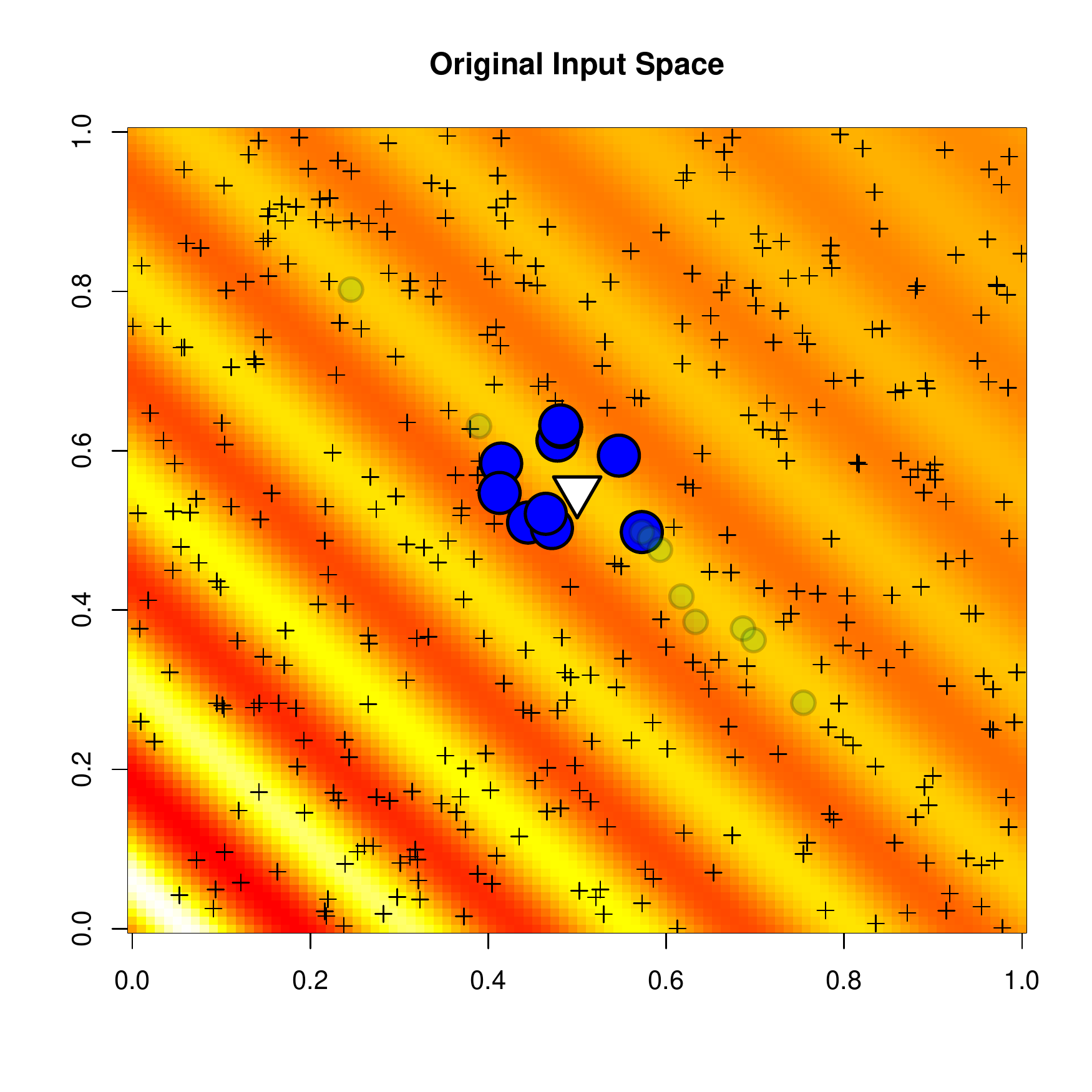}}
    \raisebox{-0.5\height}{\includegraphics[scale=0.28]{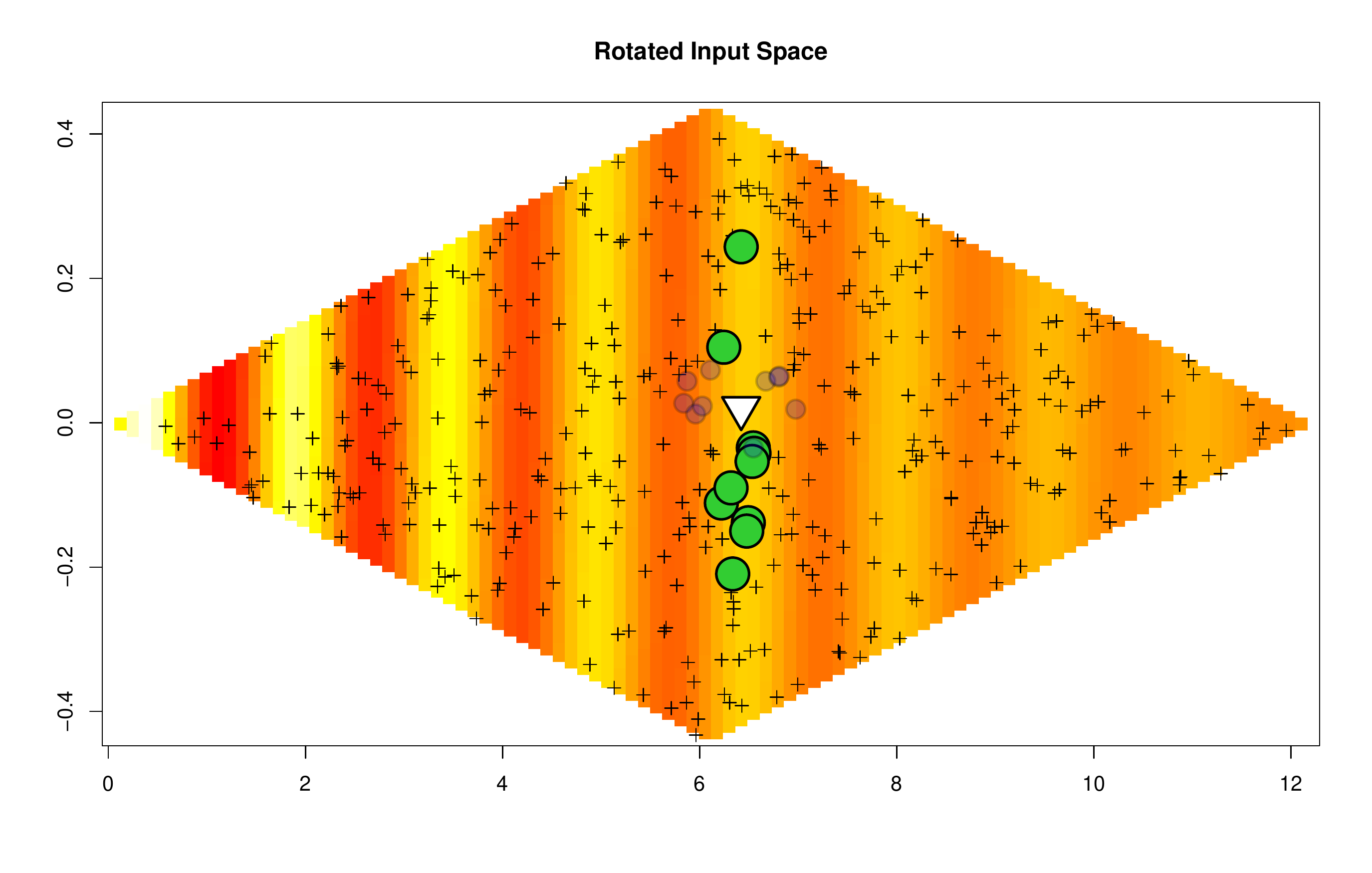}}
    \vspace{-0.5cm}
    \caption{The function $f(x) = \sin(x+y)\cos(x+y)e^{-\frac{x+y}{10}}$ with $x,y$ varying from $-2\pi$ to $2\pi$ rescaled to $[0,1]$, before (left) and after (right) $\LL_{\nu_l}$ rotation
    In both panels, the black + represent the training set and solid circles represent the 10 nearest points to an arbitrary prediction location, itself represented by the large white triangle. Faded circles give nearest neighbors from the other plot. Note that the rotated plot is not to scale for ease of viewing.}
    \label{fig:neighbors}
\end{figure}

We shall now visually explore the effect preprocessing can have on the sets of NN. Specifically, points which are farther from the prediction location along axes with little influence, but closer along axes with much influence, are comparatively favored.
Figure \ref{fig:neighbors} illustrates this principle, revisiting the ridge function of Figure \ref{fig:ridge}. 
In this toy example, we sample $400$ input locations uniformly at random in the 2d input domain, then apply Lebesgue-measure prewarping.
The left panel  shows the original input space, while the right plot shows the new input space after applying a $\LL_{l}$ rotation. 
The training set (black +'s) and prediction location (white triangle) are the same in both, but the closest points (solid circles) are changed.
In each panel, the faded circles give the locations of the solid circles from the other plot. We can see that the response value at the ten nearest neighbors is much closer to the value at the predictive location after the warping (right) than it is before (left).
%If the next step after preprocessing were to conduct NN prediction, we can visually verify that the prediction with rotation will be more accurate without. 
% on second thought, I don't know how important making note that the GP learned a full rank transform (and probably always will) is really.
%\change{As can be seen from the figure, the function} {\em (Nate, do you mean the 2d function from Section 2.1?)} varies only in one direction, so if we were to have used its true but unknown active subspace matrix as a basis for rotation, the new design space would be 1 dimensional. However, the GP based posterior estimate of $\C$ still gives a two dimensional transformation as the model has not completely ruled out a full rank $\C$.

%We have described above how rotation can aide local models insofar as it yields a globally isotropic function in expectation. Besides this, a key component of local modeling is neighborhood selection, the process of choosing which training points should be included in a specific local model. Arguably the simplest means of forming a neighborhood of size $K$ is to select those $K$ points nearest to the point of interest in terms of Euclidean distance. This is another channel through which linear preprocessing can increase performance of the local model, as performing neighborhood selection after rotation incorporates information from the sensitivity analysis. 

\section{Numerical Experiments} \label{sec:numerical}

We shall now present results of experiments devised to quantitatively evaluate sensitivity prewarping in predictive exercises.  We begin with outlining the comparators and metrics, followed by implementation details, and the actual experiments.  {\sf R} scripts reproducing all figures shown in this document may be found here: 
\url{https://github.com/NathanWycoff/SensitivityPrewarping}
%\url{https://github.com/anonymized/anonymized}

\subsection{Implementation details, comparators and metrics}
\label{sec:numimp}

%We conduct these via Monte Carlo, grading premodeling techniques based on the accuracy of downstream local models. 
The preprocessing methods will be assessed based on their effect on the performance of downstream local models. % in terms of out of sample Mean Squared Error as well as out of sample Score, estimated via randomly sampling test locations.
%As baselines, we will be using a Gaussian process regression simply fit on some random subset of the data (\texttt{sGP}) and laGP without preprocessing (\texttt{aGP}).
As baselines, we entertain GPs fit on random data subsets, which we'll denote \texttt{sGP}, as well as $k$-NN (\texttt{KNN}), laGP (\texttt{laGP}), and the Vecchia approximation (\texttt{vecc}) on the full, original dataset. 
Implementations are provided by {\sf R} packages \texttt{hetGP }\citep{hetGP,Binois2018practical}, \texttt{FNN} \citep{FNN}, \texttt{laGP} \citep{laGP,Gramacy2015lagp}, and \texttt{GpGp} \citep{Guinness2018,GpGp}, respectively.
These will be compared to \texttt{KNN}, \texttt{laGP} and \texttt{vecc} with the four specific prewarping methods proposed in Section \ref{sec:methtrans}.
The Bandwidth Scaling $\LL_{\mathrm{ARD}}$ will be denoted by prefix \texttt{B}, Lebesgue-measure prewarping $\LL_{l}$ by prefix \texttt{L}, sample-measure prewarping $\LL_{s}$ by \texttt{S}, and the range sensitivity prewarping by \texttt{R}.  
Further, we will consider truncation for all four prewarping techniques which is denoted by a postfix of $\texttt{T}$. 
%Thus, $\texttt{S-laGP}$ represents the performance of fitting an laGP model using Algorithm \ref{alg:asrot} with the sample measure but without any truncation (as that would be represented by $\texttt{S-laGP-T}$).

For each test function, we first generate data using either a random Latin Hypercube Sample \citep[LHS;][]{Stein1987} via the {\sf R} package \texttt{lhs} \citep{lhs} for synthetic data,  or via uniform random subsampling with existing/observational data, which we then randomly split into train and test sets. 
Then, we fit the baseline models for $\y$ given $\X$ and calculated their performance.
Next, we conducted the sensitivity analyses using $5$ subsamples each of size 1,500 in all experiments, using GP regression to estimate kernel hyperparameters, as well as the nugget term, via MLE \citep{Gramacy2012b}.
Afterwards, we compute the associated transformations to warp each $\X$, yielding each $\Xtrans$, as outlined in Algorithms \ref{alg:bandscale} and \ref{alg:asrot}.
Finally, each local model is fit to $\Xtrans$ versus $\y$ for each  $\Xtrans$ created by the different transformations, and their performance on each recorded. This process is repeated for 10 Monte Carlo iterations.

In surrogate modeling, quantification of uncertainty is often high priority, so we define performance using not only the Mean Square prediction Error (MSE), but also logarithmic Score \citep{Gneiting2007}.
For GP predictors, this is defined as the log likelihood of the response at a prediction location given the predictive mean and variance at that point using our assumption of Gaussianity for the response \citep[Eq.~25]{Gneiting2007}.  Since NN is typically not deployed in situations where uncertainty quantification is desired, we omit score calculations for it. 

%From a practical perspective, while the above decomposition suggests a change of basis, it does not provide guidance on how many dimensions to truncate. We simply select truncation dimension via cross validation, which is a common approach to scalar complexity parameters in variable selection contexts, such as the lasso.  Details are provided in Section \ref{sec:numerical} alongside other implementation considerations.  {\em (Nate, this is a reminder to dump implementationdetails like this into it's own sub-section, perhaps at the start of the empirical work in Section 4.)}  

%laGP was fit using the Active Learning Cohn \cite{Cohn1994} neighborhood selection technique with the default neighborhood size of 50.
%We will also consider truncation after each of these transformations, which will be denoted by a $\texttt{T}$ suffix.
%The Vecchia approximation is deployed in the exact same scenarios using the default conditioning set size of 30. For KNN regression, we also used the default $K=3$.   

%We used the R packages \texttt{laGP} to do local Gaussian processes using the default arguments, \texttt{activegp} to estimate the active subspace, \texttt{GpGp} to fit the Vecchia models, and \texttt{FNN} for KNN regresssion.    {\em (Nate, see comments on proper cites above.)} 

While calculation of $\C$ can involve sophisticated machinery, we have endeavored to make its application as simple as possible. With the {\sf R} package \texttt{activegp} \citep{activegp,sasl} loaded, prewarping is as straightforward as:
\begin{verbatim}
  R> Lt <- Lt_GP(X, y, measure = "lebesgue")  ## or measure = "sample"
  R> Z <- X %*% Lt[,1:r]                   ## r is truncated dimension
\end{verbatim}
%where \verb!"lebesgue"! may be replaced by \verb!"sample"! and \texttt{r} indicates the truncated dimension, following Algorithm \ref{alg:trunc}.

\subsection{Observational Data}

%\emph{!Put the retrieved url in the bib file, maybe the @note part?!}
We first consider two high dimensional observational datasets. 
The Communities and Crime dataset \citep{Redmond2002} combines census and law enforcement statistics from the United States.
The task is to predict crime rate per capita given 122 socio-economic indicators measured on 1,994 individuals.
The Temperature dataset \citep{Cawley2006} involves temperature forecasting given the output of a weather model, and consists of 7,117 observations and 106 features.% output by the weather model. 

\begin{figure}[ht!]\textbf{}
    \centering
    \includegraphics[scale=0.34,trim={0 3em 0 0},clip]{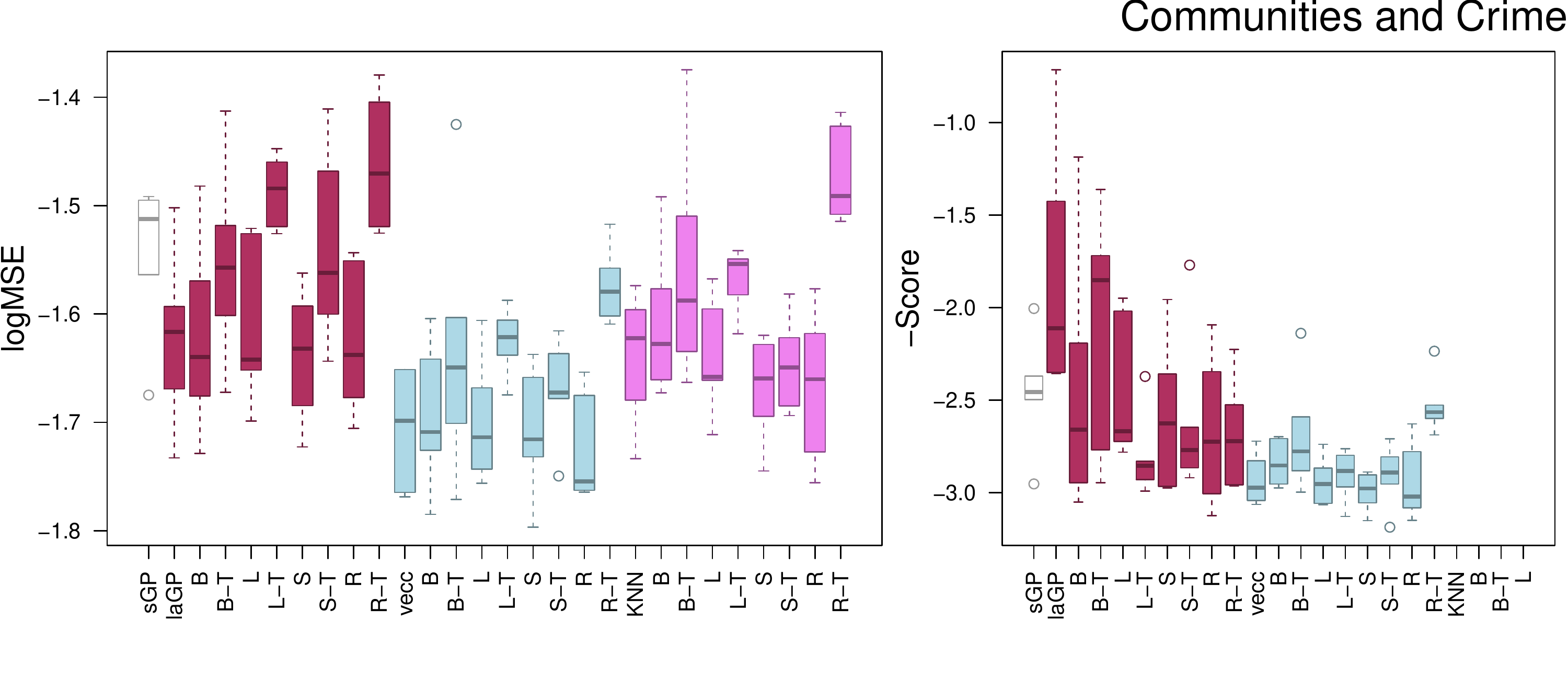}
    \includegraphics[scale=0.34,trim={0 3em 0.6em 0},clip]{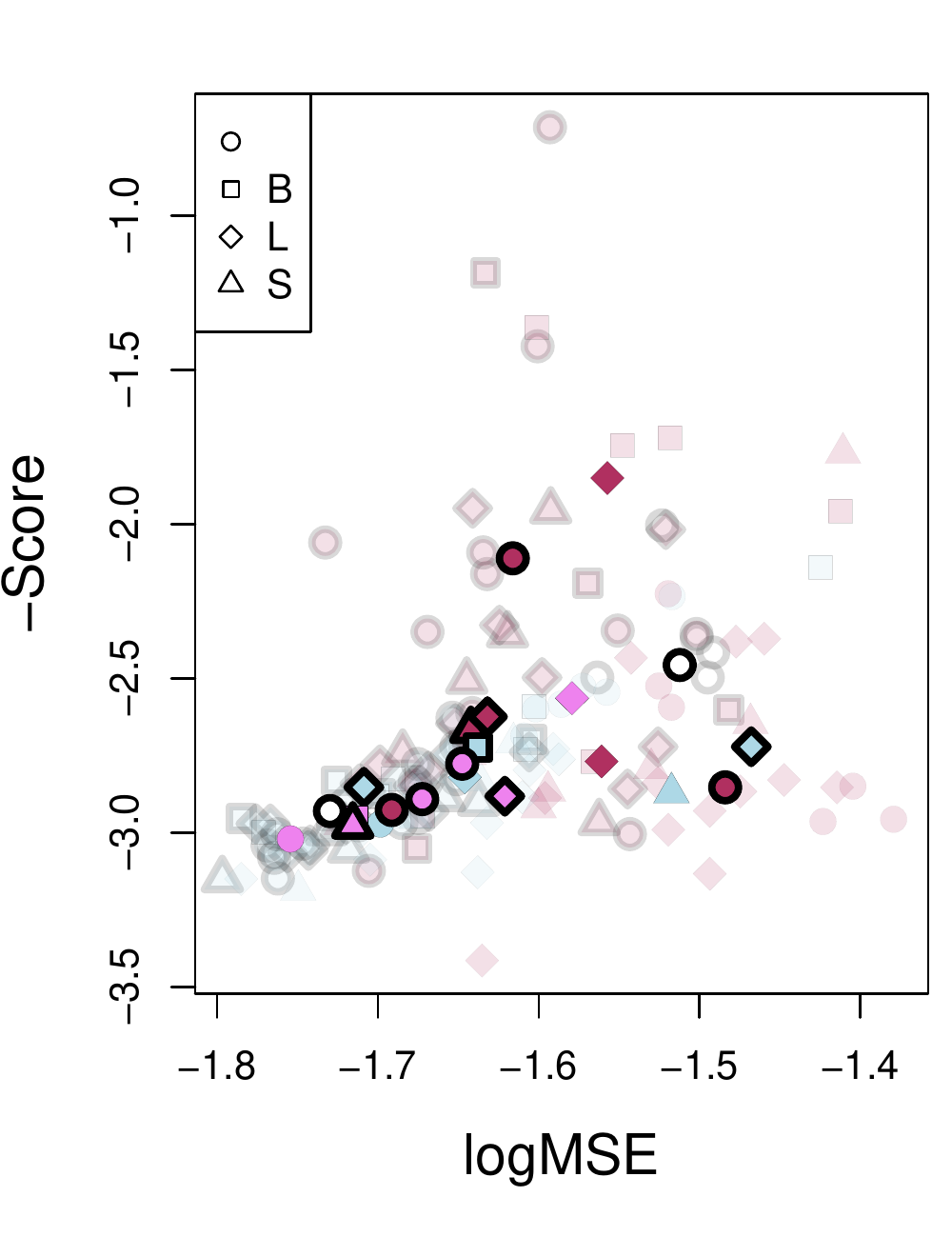}
    
    \vspace{0.25cm}
    
    \includegraphics[scale=0.35,trim={0 3em 0 0},clip]{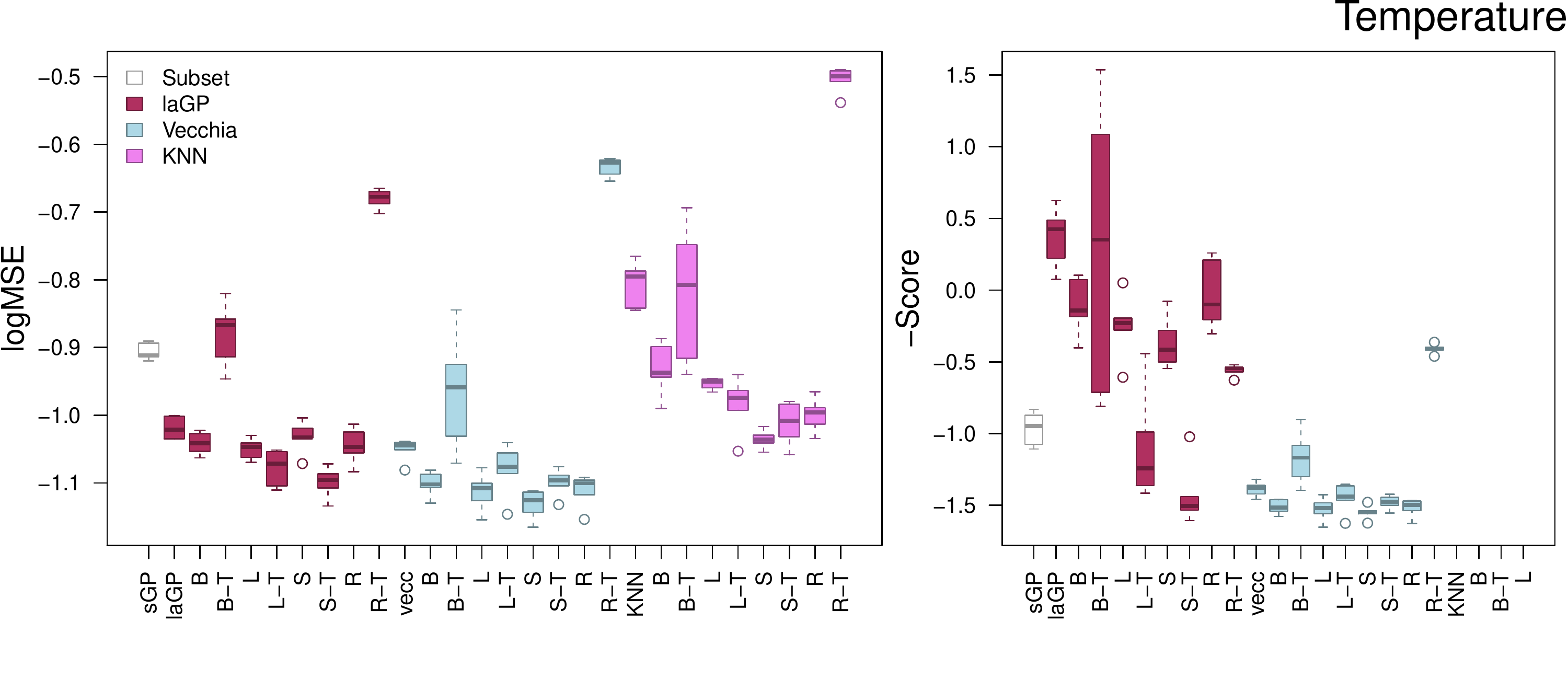}
    \includegraphics[scale=0.35,trim={0 3em 0.6em 0},clip]{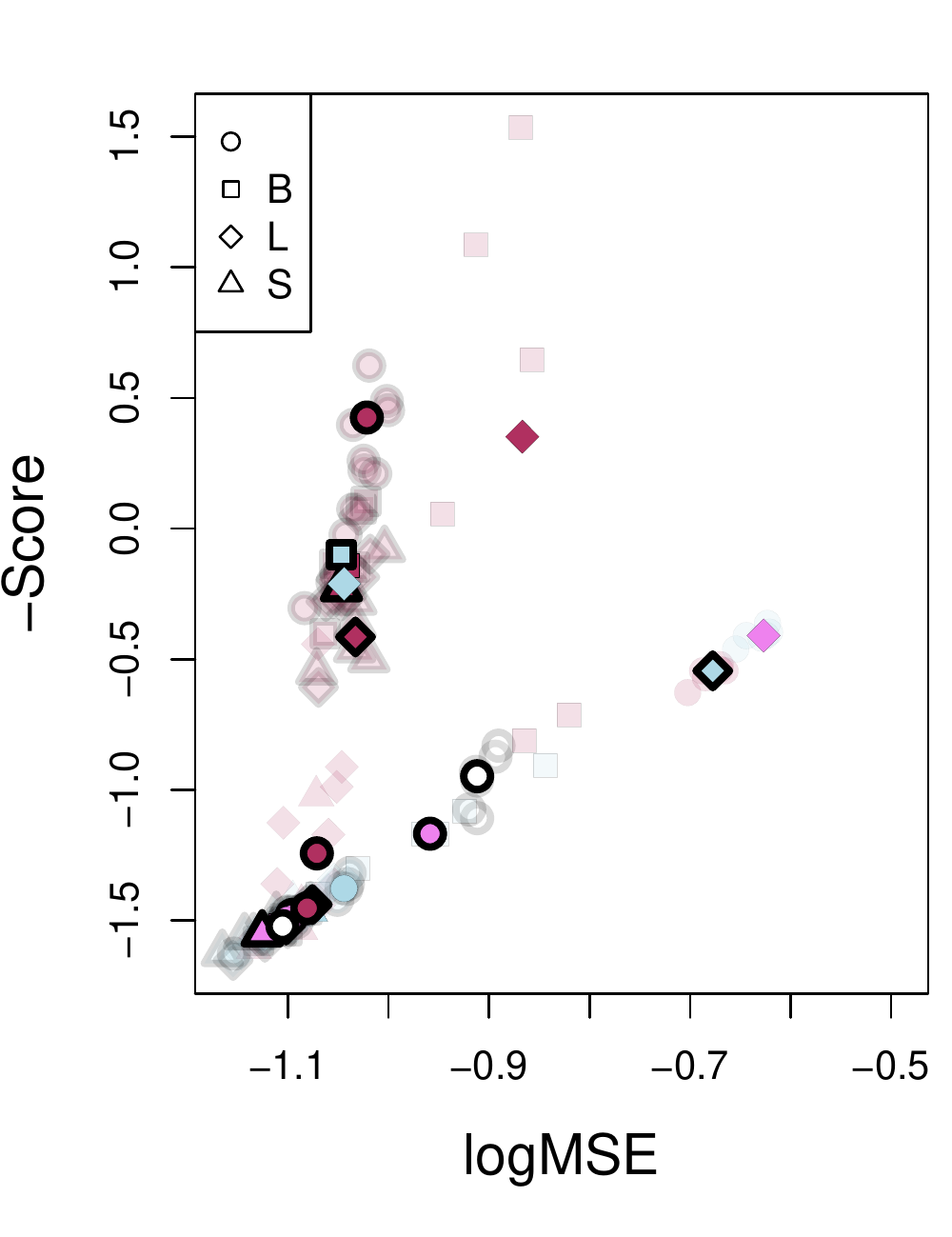}
    \caption{Results on two observational test problems. {\em Left and Center:} the $y$-axis gives either $\log_{10}$ MSE or negative Score (smaller is better). The letter before the name, \texttt{B}, \texttt{L} and \texttt{S}  represents the transformation used for prewarping (if there is one); \texttt{T} denotes truncation. \textbf{Bold} names indicate prewarping. Models that failed to fit are left blank. {\em Right:} logMSE vs -Score for each run; faded icons indicated individual run while solid icons give group medians. Circles indicate no prewarping, solid borders indicate no truncation.}
    \label{fig:obsbox}
\end{figure}

The performance of the competing methods is given in Figure \ref{fig:obsbox}. We find that truncation is helpful for high dimensional problems, particularly on the Temperature dataset, and more so for the active subspace rotations than for the axis scaling methods (Bandwidth and Range). We also find that the $\LL_{s}$ generally outperforms $\LL_{l}$. This is because the observational data are not uniformly distributed, which has two implications. First, since the training set is not uniformly distributed, Sample measure overemphasizes certain parts of the input space compared to Lebesgue. Second, because the test set was formed by random sampling, these same parts of the input space that we have implicitly tuned our $\LL$ estimate to are those parts of the input space in which we tend to find testing locations.
In other words, there is simply a mismatch between the probability distribution from which the observational data were drawn and that with respect to which $\LL_{l}$ is defined. 
We see that the preprocessing differentiated itself the least on the Communities and Crime problem, potentially because this problem consisted of significantly fewer observations, at around $1{,}000$, making it difficult to estimate the rotation, and leading to high variance.

\begin{figure}[ht!]
    \centering
    \includegraphics[scale=0.46]{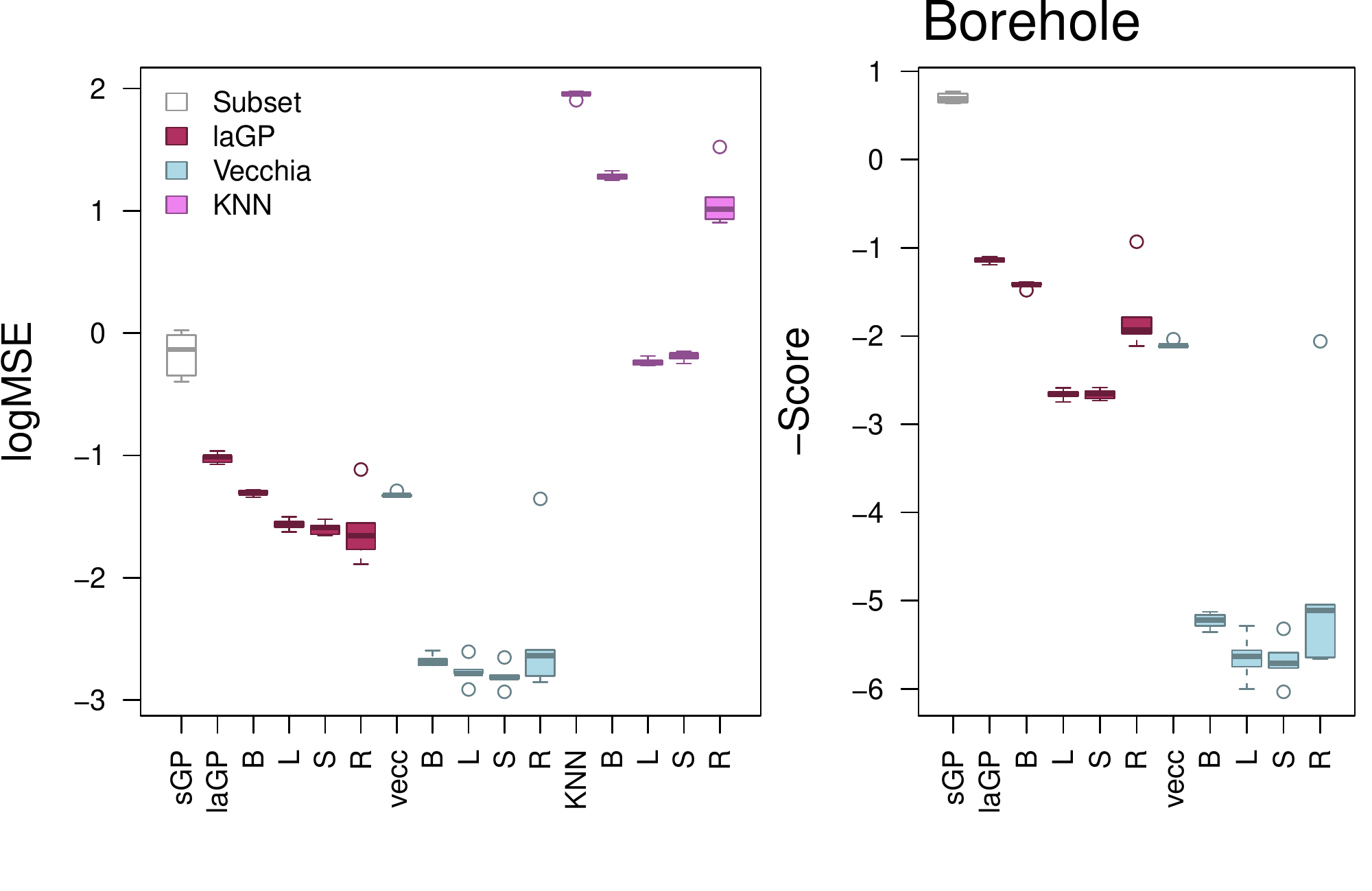}
    \includegraphics[scale=0.46]{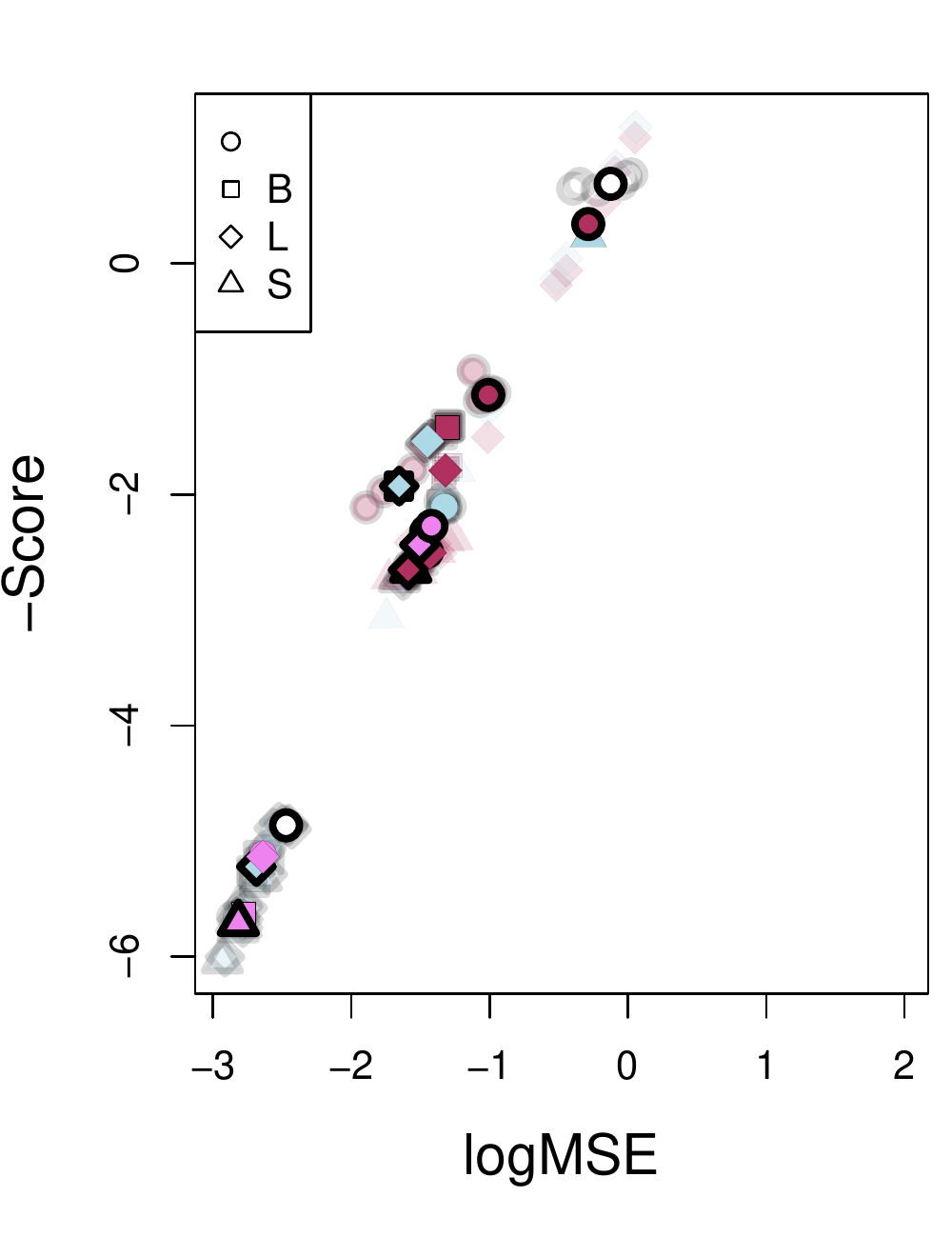}
    \includegraphics[scale=0.46]{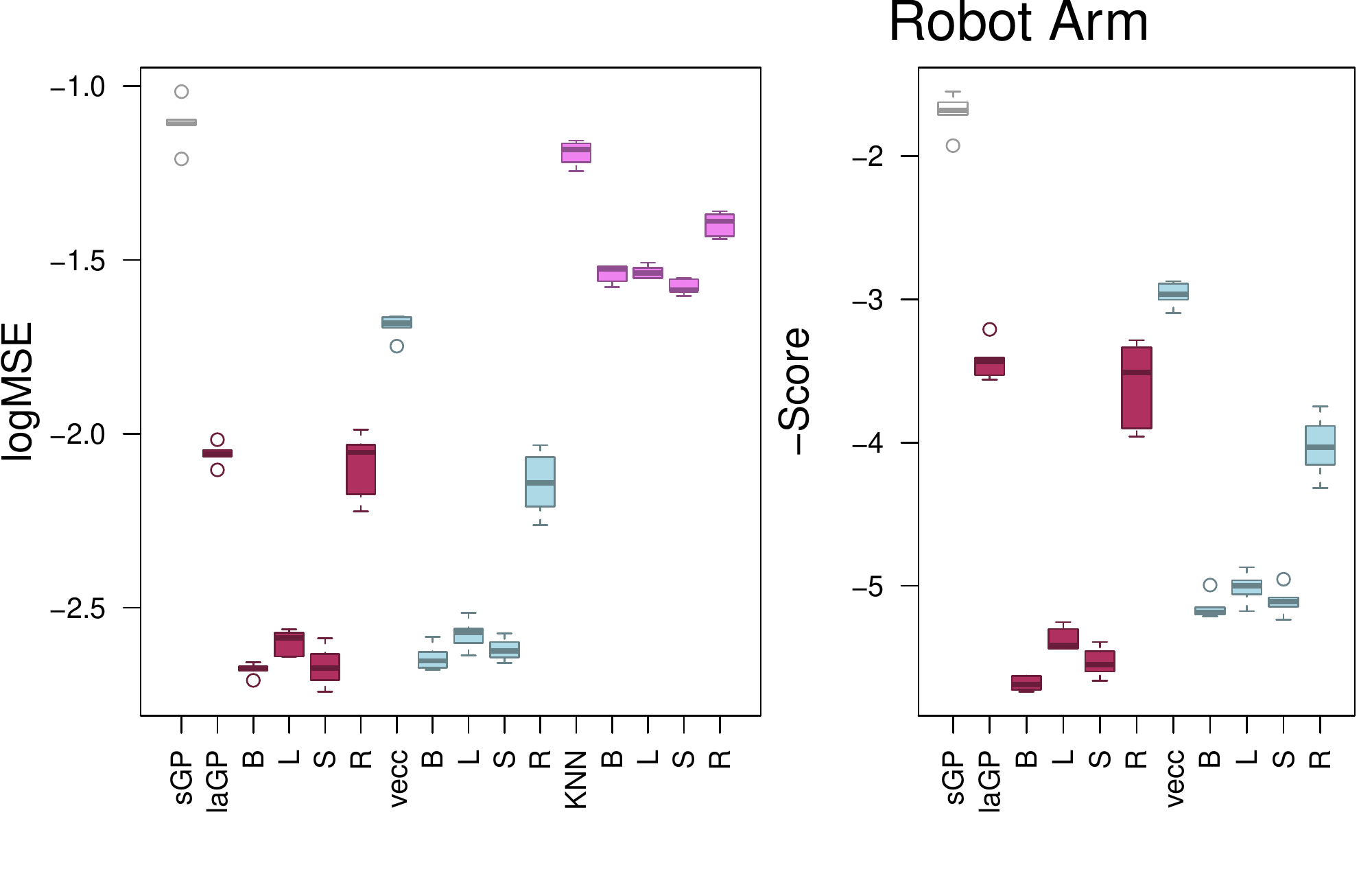}
    \includegraphics[scale=0.46]{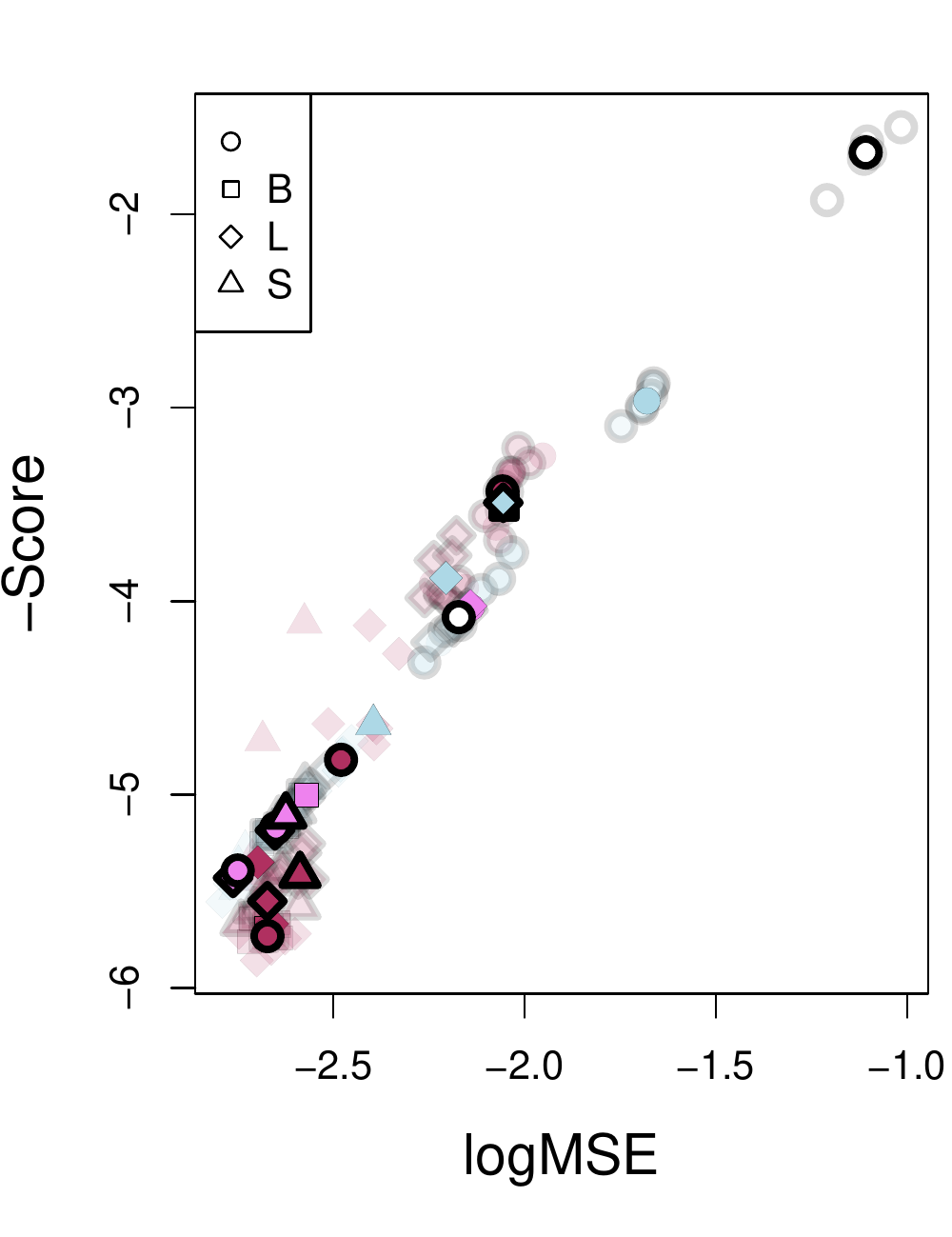}
    \includegraphics[scale=0.46]{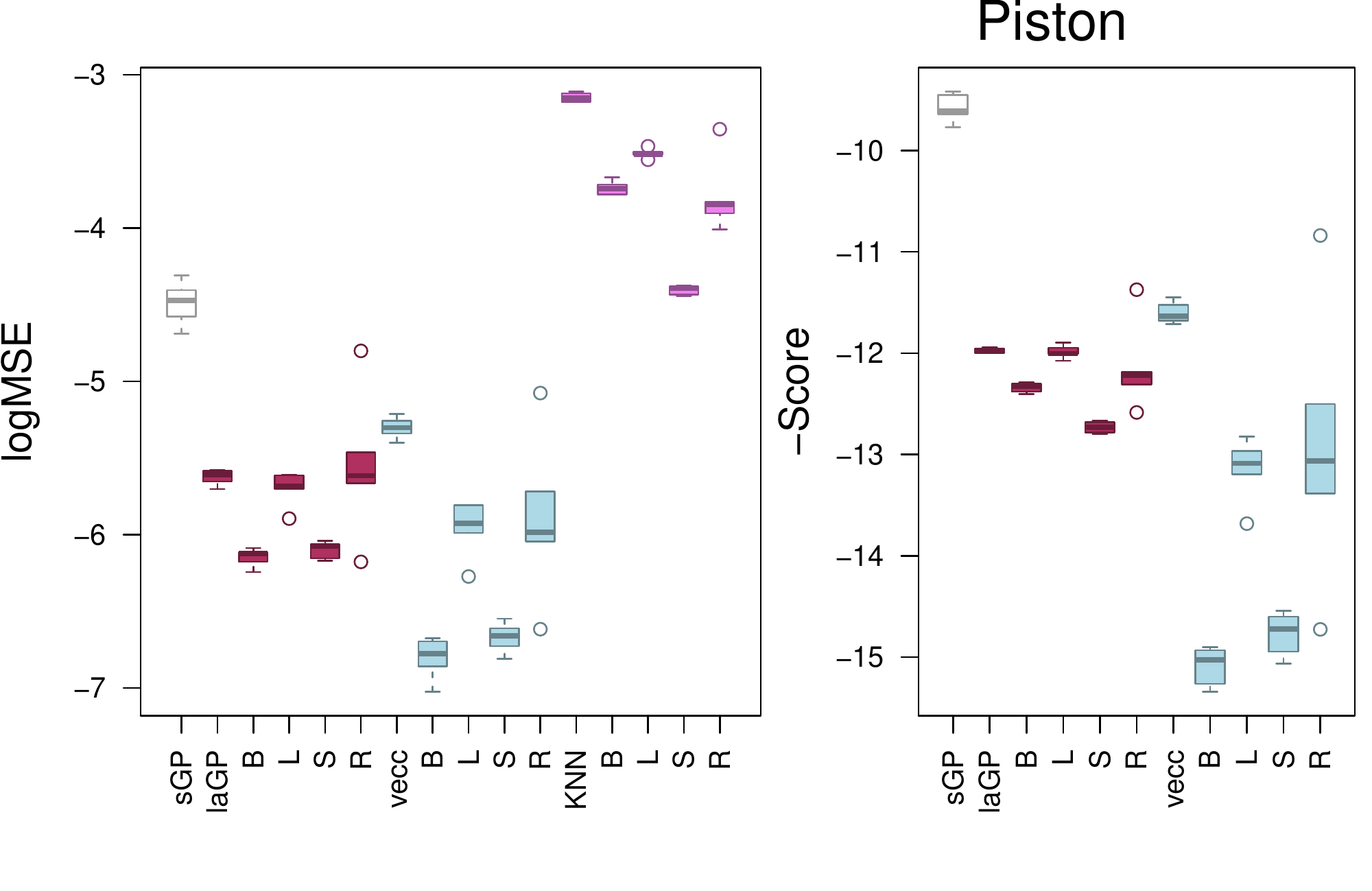}
    \includegraphics[scale=0.46]{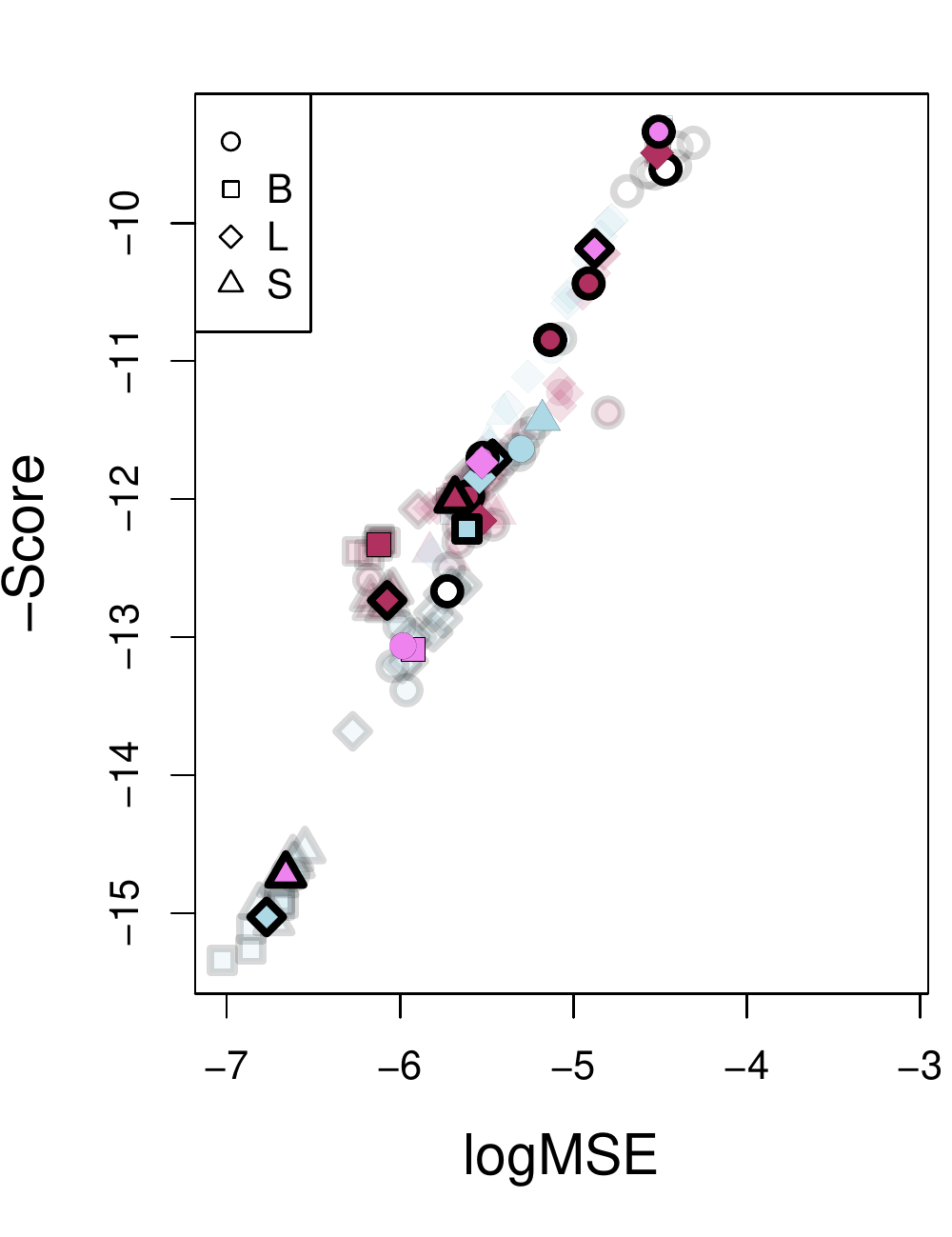}
    \caption{A comparison on common test functions with $n=40{,}000$ runs. See Figure \ref{fig:obsbox} caption.}
    \label{fig:lowd}
\end{figure}

\subsection{Benchmark Test Functions}

We next evaluated the proposed methodology on benchmark test functions \citep{simulationlib} where we found that prewarping increased performance in terms of both MSE and Score.
In particular, we ran the competing methods on the Borehole \citep[$\nvar=8$]{Harper1983}, Robot Arm \citep[$\nvar=8$]{An2001}, and Piston \citep[$\nvar=7$]{Ron1998} functions with a training set size of $40{,}000$ and test set size of $2{,}000$ for each, sampled from a random LHS. 

The results, shown in Figure \ref{fig:lowd}, indicate that prewarping can be quite beneficial for local modeling in terms of predictive accuracy.
On these low dimensional problems, each method performed similarly regardless of whether truncation was applied, so we have omitted truncation in the results.
On all three problems, all forms of prewarping greatly outperform respective baselines.
On the Borehole problem the AS based methods $\LL_{l}$ and $\LL_{s}$ outperform both the baselines and $\LL_{\mathrm{ARD}}$ in terms of both MSE and Score. The Range prewarping seems to have a slight edge in MSE and a slight disadvantage in Score. 
On the Robot Arm function, we find that all prewarping methods are pretty similar, with the sample-measure $\LL_{s}$ generally having a slight edge. The Range transformation seems to be at a disadvantage on this problem.
Finally, on the Piston problem, prewarping generally leads to a decrease in MSE, though which particular method is ahead depends on the local model considered. 
Range again does about the same as no prewarping.

\subsection{The Jones MOPTA Problem}

\begin{figure}
    \centering
    \includegraphics[scale=0.375,trim=10 0 29 20]{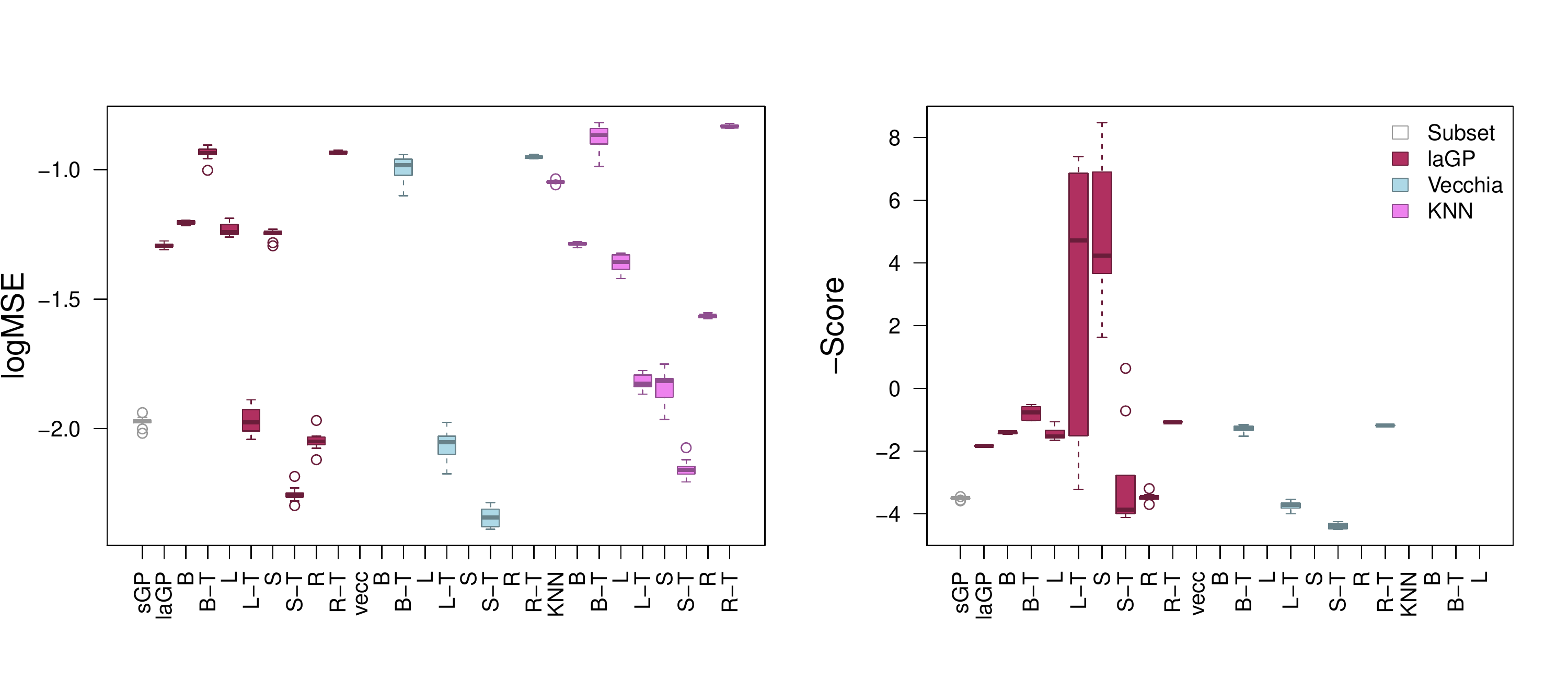}
    \includegraphics[scale=0.35,trim=0 0 20 0]{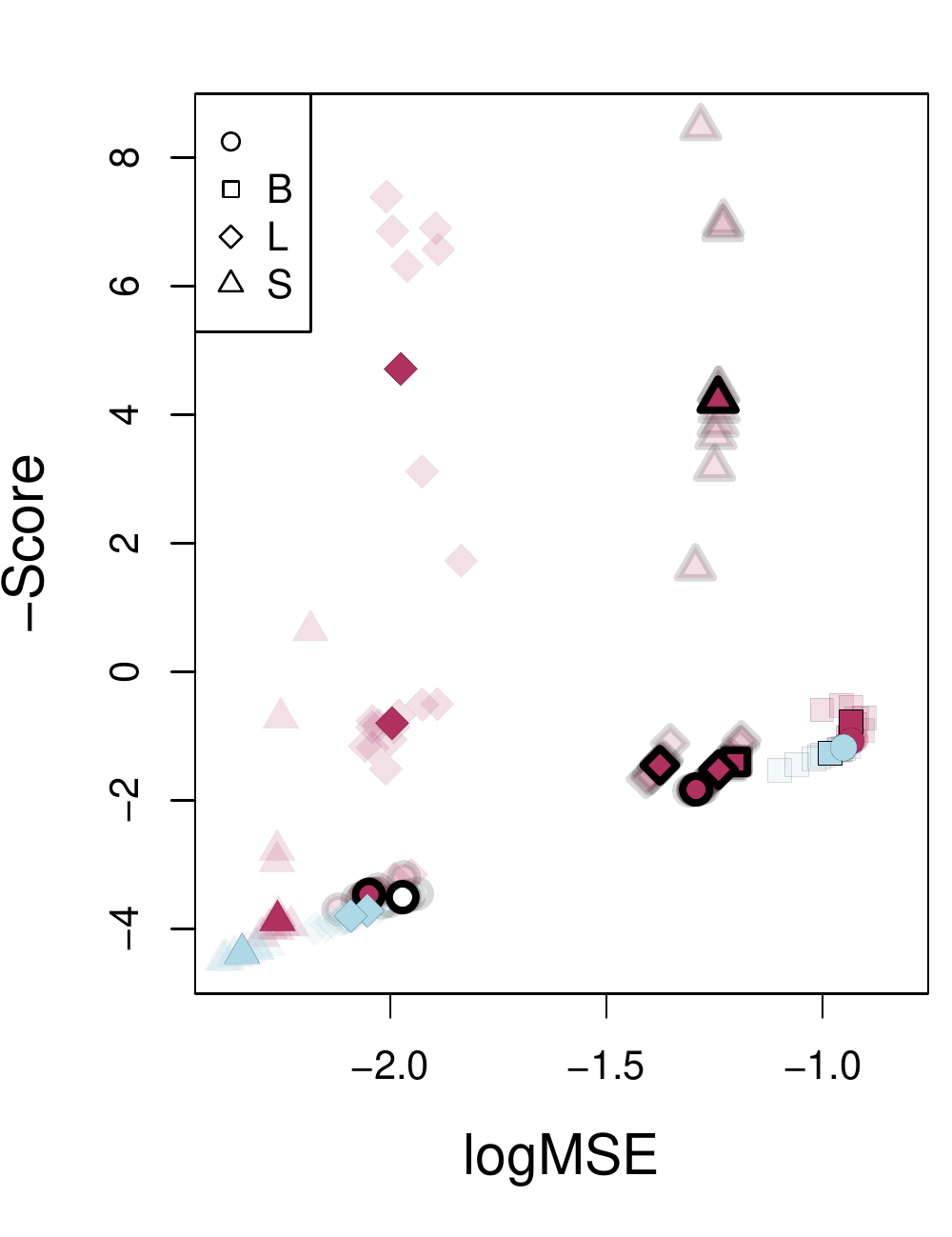}
    \caption{A comparison on the 124d MOPTA function. See Figure \ref{fig:obsbox} caption.}
    \label{fig:mopta}
\end{figure}

In this section, we study the performance of prewarping on an optimization problem presented by General Motors at the 2008 ``Modeling and Optimization: Theory and Applications (MOPTA)'' conference \citep{jones2008large}.
%This problem concerns product automotive design via computer experiments. 
The input variables characterize the design of the automobile, such as materials, part gauges, and shape, which determine the results of several crash simulations.
The problem is to minimize the mass of the configuration, while observing constraints, such as the durability of the vehicle and the harshness of the crash.
This is a constrained optimization problem involving 124 input variables and 68 constraints.
While the standard approaches to smooth, high dimensional, constrained optimization are primarily gradient-based, the simulator, a multi-disciplinary effort, does not provide gradients with respect to inputs, and numerical noise means finite differencing approaches are not applicable. 

Various authors have proposed sophisticated solutions for this challenging problem, including those based on Bayesian optimization, evolutionary strategies, or both.
\cite{Regis2011} proposed fitting a surrogate model to each constraint as well as the objective function to launch a stochastic search for good feasible solutions. 
\cite{Beaucaire2019} tackled the optimization problem by effectively using an ensemble of surrogates, while \cite{Regis2012} combined surrogate modeling approaches with evolutionary algorithms, and \cite{Regis2017} combined surrogate modeling with trust region methods.
However, this article is concerned with the large data regime, which is generally not the case when conducting Bayesian optimization. 
To study Jones MOPTA as an emulation problem, we simply treat the sum of the objective and all of the constraints as a black-box function to approximate. 
This black-box is of interest as such augmented objective functions form the basis of penalty-based approaches to constrained optimization \citep{Nocedal2006}.

We sampled $500{,}000$ points uniformly at random in the input space, treating $2{,}000$ as a test set, chosen randomly.
%\texttt{vecc} ran into numerical issues when fit directly on the high dimensional space, so we have only retained truncated versions of it.
We chose not to include \texttt{vecc} fit on the untruncated data as the runtime was too long.
As the results in Figure \ref{fig:mopta} show, in terms of MSE,
prewarping without truncation can somewhat improve performance, but throwing in truncation as well results in improvements of an order of magnitude or more using doing AS prewarping ($\LL_{s}$ or $\LL_{l}$). 
The exception is \texttt{S-KNN}, which is able to achieve competitive accuracy without truncation.
In terms of score, it would appear that prewarping without truncation can result in a significant decrease in performance compared to baseline.
Indeed, looking at the scatterplot (Figure \ref{fig:mopta}, right), we see that without truncation, the various local models and prewarpings form a spectrum of solutions trading MSE for Score, whereas the truncated AS prewarped local models significantly outperforms in terms of both. However, this trend is not universal among prewarpings: the Range prewarping performs very well in terms of MSE without truncation, but not with. It seems as though the Range prewarping can offer a good warping of the space, but not one amenable to truncation. 
% Notably, which prewarping is used has significantly greater impact than which local model is used.

%\subsection{``Off-Policy" Training}
%
%If we're tied in random designs and losing in observation designs, why do $C_{\nu_l}$ at all? We need an example where the distribution of observed $\X$ and the prediction locations are different. For instance, choose design via EI or something (would be even better if it was active subspace sequential). Space-filling probably won't cut it. 

\section{Conclusions and Future Work}\label{sec:conclusion}

We introduced Sensitivity Prewarping, a simple-to-deploy framework for local surrogate modeling of computer experiments.
Specifically, we proposed the heuristic of warping the space such that a global sensitivity analysis would reveal that all directions are equally important, and showed specific algorithms based on the ARD principle and/or AS to achieve this.
%We showed mathematically that our proposed procedure induces a regression problem with no global structure as measured by the Active Subspace matrix\change{, and proposed this as . 
By learning directions of global importance, we free each of the local models from individually learning global trends, and instead allow them to focus on their prediction region. Our prewarping effectively defines a new notion of distance which has the dual benefit of improving both neighborhood selection and the value of distance in prediction. We also proposed a subbagging procedure for scaling up inference of the AS as estimated via a GP. 

Generally, our numerical experiments revealed that prewarping yields significant benefits in terms of predictive accuracy, as measured by MSE, as well as predictive uncertainty, as measured by Score. We showed how rotations can improve inference on low dimensional test functions, and how truncation can be transformative in high dimensional problems.  
Given the ease of implementation and the important improvement in predictive accuracy, we submit that this procedure has broad applicability. 

We focused on three specific sensitivity analyses and three specific local models, but there is plenty of room for further inquiry.
Deploying this framework with nonlinear sensitivity analysis (i.e., that which can measure the importance of nonlinear functions of the inputs) could be fruitful, for instance with Active Manifolds \citep{Bridges2019}. 
It would also be interesting to study what sensitivity techniques could be expected to perform well when paired with a given local model.

%In Section \ref{sec:bgbign}, we reviewed many approaches to scaling GP processes to large datasets, 
%Room for further work can also be found in scaling up estimation of $\C$. %It is possible that our subbagging approach could be improved upon
Another area where future work could lend improvements is in large scale estimation of $\C$. In this article, we proposed a subbagging solution, but many other approaches are conceivable. For instance, $\C$ could be computed by using existing approximations to the kernel matrix, such as the Vecchia approximation. An alternative would be to deploy Krylov subspace methods, which have shown great promise in scaling GPs \citep{Wahba1995,Gibbs97,Pleiss2018,Dong2017}, to develop stochastic algorithms either to estimate the matrix $\C$ itself or its leading eigenspace directly \citep{Golub2010}.

%GP modeling with gradient-based hyperparameter inference (for $\theta$), via maximization or posterior sampling, boils down to repeated computation of $\K^{-1}\y$, $\textrm{det}(\K)$, and $\textrm{tr}(\K^{-1} \frac{\partial \K}{\partial \theta})$ for all $\theta$ entertained. So far, the methods we have discussed may be viewed as approximate models for which computation of these quantities is possible. Another approach is to use the full GP model, but instead approximate the computations of those quantities. Various authors  have proposed to use techniques from the cutting edge of computational linear algebra to do so. In particular, the linear system $\K^{-1}\y$ may be solved via Krylov subspace methods, which rather than computing a decomposition of $\K$ instead computes the action of $\K$ on $\y$, and then on the resulting vector $\K \y$, and then on $\K^2 \y$, and so on\footnote{Here, $\mathbf{A}^2 = \mathbf{A}\mathbf{A}$, not the elementwise square}. The space spanned by $\{\y, \K\y, \K^2\y, \ldots, \K^{l-1}\y\}$ is called the \textit{Krylov subspace} of dimension $l$. Happily, with Krylov subspace vectors in hand, the linear algebraic quantities needed for GP training and inference may be estimated efficiently through randomized algorithms\cite{Golub2010}.   {\em (Nate, a downside to this paragraph is that it -- I think -- isn't germane to your contribution.  It might be long in the discussion section instead.)}

Arguably, the weakest link of this approach is the GP fit in the first stage which produces our estimator of $\C$, required to compute $\LL$ in the AS approach.
This is because local models can compensate for breaches of our GP assumptions such as stationarity and homoskedasticity, while the global fit cannot.
Hence, designing techniques for estimation of $\C$ via more sophisticated models is likely to be a fruitful thread of research.
Deep GPs \citep{Damianou2013} are a natural next step, and have been recently studied in the context of computer experiments \citep{Sauer2020}.
Finally, the simulators we studied in this article all accepted a vector of inputs and returned a scalar response. Extensions to vector-valued, discrete, or functional responses would increase the breadth of problems this framework can take on.

\bibliographystyle{chicago}
\bibliography{main}

\end{document}